\documentclass{article}

\usepackage{algorithm}
\usepackage[noend]{algpseudocode} 

\usepackage{tikz}
\usepackage{amsfonts}
\usepackage{amsmath, amssymb}
\usepackage{booktabs}
\usepackage{multirow}
\usepackage{changepage}
\usepackage{ctable} 
\usepackage{caption}
\usepackage{subcaption}
\usepackage{calc}     
\usepackage{makecell} 
\usepackage[affil-it]{authblk}

 \usepackage{etoolbox}
\usepackage{natbib}
 \bibpunct[, ]{(}{)}{,}{a}{}{,}%
\title{Deep Learning--Accelerated Multi-Start Large Neighborhood Search for Real-time Freight Bundling}

\author[1]{Haohui Zhang}
\author[1]{Wouter van Heeswijk}
\author[1]{Xinyu Hu}
\author[2]{Neil Yorke-Smith}
\author[1]{Martijn Mes}

\affil[1]{Department of High-Tech Business and Entrepreneurship, University of Twente, Enschede, Netherlands}
\affil[2]{STAR Lab, Delft University of Technology, Delft, Netherlands}

\date{\today}

\begin{document}

\maketitle

\begin{abstract}
Online Freight Exchange Systems (OFEX) play a crucial role in modern freight
logistics by facilitating real-time matching between shippers and carrier.
However, efficient combinatorial bundling of transporation jobs remains a
bottleneck. We model the OFEX combinatorial bundling problem as a
multi-commodity one-to-one pickup-and-delivery selective traveling salesperson
problem (m1-PDSTSP), which optimizes revenue-driven freight bundling under
capacity, precedence, and route-length constraints.
The key challenge is to couple combinatorial
bundle selection with pickup-and-delivery routing under sub-second latency. We
propose a learning--accelerated hybrid search pipeline that pairs a Transformer Neural
Network-based constructive policy with an innovative Multi-Start Large
Neighborhood Search (MSLNS) metaheuristic within a rolling-horizon scheme in
which the platform repeatedly freezes the current marketplace into a static
snapshot and solves it under a short time budget. This pairing leverages the
low-latency, high-quality inference of the learning-based constructor alongside
the robustness of improvement search; the multi-start design and plausible seeds
help LNS to explore the solution space more efficiently. Across
benchmarks, our method outperforms state-of-the-art neural combinatorial optimization
and metaheuristic baselines in solution quality with comparable time, achieving
an optimality gap of less than 2\% in total revenue relative to the best
available exact baseline method. To our knowledge, this is the first work to
establish that a Deep Neural Network-based constructor can reliably provide
high-quality seeds for (multi-start) improvement heuristics, with applicability beyond the \textit{m1-PDSTSP} to a broad class of selective traveling salesperson problems and pickup and delivery problems.
\end{abstract}
\section{Introduction}\label{sec:Intro}

Trucking is the backbone of freight transportation in many countries. In recent
years, Online Freight Exchange Systems (OFEX) platforms (e.g., Trans.eu and
Uturn in Europe; Full Truck Alliance in China; Convoy, Coyote Logistics and Uber
Freight in the United States) have become a common marketplace for shippers and
carriers to connect and arrange services \citep{fazi2025freight}. These digital
platforms facilitate seamless freight matching, real-time monitoring, and
payment processing between shippers and carriers. By lowering search and
transaction costs, these platforms assist in shippers reducing transport
expenses and enable carriers, especially small operators, to find profitable
backhauls, thereby reducing empty kilometers and improving market efficiency.

Beyond functioning as digital marketplaces, OFEX platforms have compelling
economic incentives to integrate sophisticated algorithms to automate freight
matching and optimize load allocation by analyzing comprehensive market data,
carrier capacities, and demand patterns. By bundling shipments and recommending
route-level assignments, these platforms can enable carriers to assemble
feasible, profitable bundles of loads, thereby increasing revenue, equipment
utilization, and ultimately driving a more efficient and profitable freight
ecosystem.


Designing freight bundling algorithms for OFEX platforms remains understudied
and highly challenging. OFEX markets are computationally centralized yet
informationally decentralized: key information (e.g., future commitments) is
private; loads and availability arrive asynchronously; and each shipment is
exclusive (awarded to at most one carrier in each lane). In practice, platforms must
recommend real-time backhaul and add-on bundles to many independent carriers
under streaming arrivals and exclusivity, creating conflicts when similar
bundles are provided to multiple carriers. Consequently, bundling couples
combinatorial selection with pickup-and-delivery routing (an NP-hard problem even
in static settings) while platforms operate under sub-second latency constraints
that preclude exact methods during live allocation. This calls for
low-latency bundling algorithms that jointly enforce pickup-and-delivery
feasibility, equipment compatibility, and carrier-level preferences within a
rolling horizon. 

We model the OFEX combinatorial bundling problem as a multi-commodity one-to-one
pickup-and-delivery selective traveling salesperson problem (\textit{m1-PDSTSP}), a
selective variant of the pickup-and-delivery traveling salesperson problem (PDTSP)
in which each request has exactly one pickup node and one delivery node;
requests are heterogeneous in size and value; and a single vehicle serves only a
subset of requests subject to capacity and route-length constraints. The
objective is to maximize revenue over all feasible pickup-and-delivery completions.

To meet real-time constraints, we adopt a rolling-horizon snapshot approach:
although the market evolves continuously, the platform repeatedly solves static
instances for a timescale shorter than market changes. To deliver high-quality
bundles rapidly, we propose a learning--accelerated hybrid search algorithmic pipeline
that combines the strengths of learning-based constructive heuristics with the robustness of improvement heuristics. Specifically, we develop the deep learning (DL)--accelerated Multi-Start
Large Neighborhood Search (MSLNS) for the \textit{m1-PDSTSP}, which leverages
the Transformer neural network \citep{vaswani2017attention} and an innovative
multi-start improvement heuristics. The Transformer-based policy scores and assembles
requests into a feasible route-level bundle, producing plausible seed solutions rapidly
that already respect pickup-and-delivery precedence, capacity, and route-length
limits. These seeds are then refined by MSLNS, which provides diversification
(multi-start + softmax-biased removal + adaptive destroy sizes) and intensification (improve with
feasibility checks) to escape local optima and reduce the remaining optimality
gap. To formalize the construction phase, we model the Transformer decoding process as a deterministic Markov Decision Process (MDP).

We elucidate the algorithmic rationale behind the hybrid pipeline and the design of MSLNS by analyzing attraction basins in neighborhood-based search, demonstrating how high-quality initializations and multi-start strategies accelerate convergence to superior local optima. Classical metaheuristics typically rely on random or greedy constructions to generate initial solutions, which often correspond to low-quality attractors located far from basins of superior local optima. As a result, substantial exploration (often requiring large neighborhood search) is needed, leading to higher computational cost and poor performance under tight latency constraints, as faced in the \textit{m1-PDSTSP}. By contrast, our neural constructor is empirically shown to rapidly produce high-quality seeds that position the search close to promising basins, enabling efficient improvement.


On benchmark instances, our approach outperforms state-of-the-art (SOTA)
learning-based constructive heuristics and metaheuristics in solution
quality and runtime: (i) Deep Neural Network (DNN)-generated solutions alone
surpass metaheuristics initialized with greedy constructions while using less
wall-clock time; (ii) appending a lightweight LNS improver to the
DNN output outperforms learning-based constructive baselines that employ
sophisticated search with less computational time; and (iii) the full DNN+MSLNS
procedure achieves the lowest optimality gap in the majority of benchmark
settings, under comparable wall-clock time. These results demonstrate the
effectiveness of proposed hybrid pipeline, its practical potential for real-time
routing, and the nontrivial role of high-quality initial solutions in improvement
heuristics.

\paragraph{Contribution.} (i) Conceptual: we formulate the \textit{m1-PDSTSP} to capture OFEX bundling and apply an attraction basin perspective to explain how initialization quality and multi-start strategies shape neighborhood-search performance. (ii) Methodological: We design an approximate feasibility-masking and reward-shaping scheme that enables efficient training of a Transformer-based policy-gradient constructor. Building on the insights from basin structure, we propose MSLNS, a multi-start large-neighborhood search procedure with softmax-biased removal and adaptive destroy sizes that exploits information embedded in learned trajectories to balance diversification and intensification.
(iii) Empirical: Through extensive computational experiments on \textit{m1-PDSTSP} benchmarks, we demonstrate that our DL--accelerated MSLNS achieves superior performance against SOTA baselines. To the best of our knowledge, this is the first systematic demonstration that a neural constructor can reliably deliver high-quality initial solutions that measurably enhance (multi-start) improvement heuristics.

The remainder of the article is organized as follows. Sec.~\ref{sec:Literature}
reviews related work. Sec.~\ref{sec:ProbDef} defines the
\textit{m1-PDSTSP}. Sec.~\ref{sec:Methodology}
describes our methodology, including the motivation of learning-seeded hybrid search and the details of the proposed DL--accelerated MSLNS.
Sec.~\ref{sec:Experiments} reports computational results and ablations.
Sec.~\ref{sec:Conclusion} concludes with a summary and directions for future
research.

\section{Related Work}\label{sec:Literature} 
We review related work on pickup and delivery problems (PDPs) and related solution approaches.


\subsection{Pickup and Delivery Problems}

Following \cite{berbeglia2007static}, PDPs can be
classified by the demand-supply structure into: (i) one-to-many-to-one (1-M-1),
where all deliveries originate at the depot and all pickups return to the
depot—also known as the TSP with pickup and delivery (TSPPD)
defined by \cite{mosheiov1994travelling}; (ii) many-to-many (M-M), in which a single commodity is
transported from supply to demand customers, yielding the one-commodity PDTSP
(1-PDTSP) defined by \cite{hernandez2003one}; and (iii) one-to-one (1-1), where each pickup is paired
with exactly one delivery, underpinning the multi-commodity one-to-one PDTSP
(m1-PDTSP) defined by \cite{hernandez2009multi}. The m1-PDTSP seeks a depot-to-depot tour
that executes all paired pickups and deliveries under vehicle-capacity limits
while minimizing travel distance.

In this study, we catalog the single-vehicle PDP formulations according to the
attributes of demand-supply structure, node, vehicle, customer and commodity
following \cite{ting2013selective}. Tab.~\ref{tab:comparison_SVRP} summarizes
representative variants and highlights their differences. Demand-supply
structure includes 1-1, 1-M-1, and M-M. Node attributes cover selectivity
(optional visits), depot supply/demand (whether the depot dispatches loads or
receives them), and multiple visits (revisits allowed). Vehicle attributes
include capacity bounds and route-length/time-window limits. Customer attributes
include time windows and service type (pickup, delivery, or both). Finally,
commodity attributes distinguish homogeneous (single-type) from heterogeneous
(multi-type) requests. 



\begin{table}[h!]
\centering
\caption{Comparison between single vehicle PDP variants and the
\textit{m1-PDSTSP}. (\textit{Sel: selectivity, Dep: depot supply/demand, Vis:
multiple visits, Ser: service, TW: time window, Cap: capacity. p: pickup only,
d: delivery only, p\&d: pickup and delivery at a visit,  p\&/d: only pickup or
only delivery or both at a visit, HO: homogeneous, HE: heterogeneous.}) We list
only one exemplary paper per problem due to space constraints.}

\setlength{\tabcolsep}{2.5pt}      
\renewcommand{\arraystretch}{1.05} 
\resizebox{0.75\columnwidth}{!}{

\begin{tabular}{ cccccccccc>{\raggedright\arraybackslash}p{10cm} } 
\specialrule{1.5pt}{.1em}{.3em} 
\multirow{2}{5em}{\textbf{Problem}} & \multirow{2}{4em}{\textbf{Scheme}} & \multicolumn{3}{c}{\textbf{Node}}& \multicolumn{2}{c}{\textbf{Vehicle}} & \multicolumn{2}{c}{\textbf{Customer}} & \multirow{2}{5em}{\textbf{Commodity}} &
\multirow{2}{5em}{\textbf{Reference}} \\
\cmidrule(lr){3-5}
\cmidrule(lr){6-7}
\cmidrule(lr){8-9}
 & & Sel & Dep & Vis & Cap & TW & TW & Ser & & \\[1pt] 
\specialrule{.5pt}{.05em}{.3em} 
STSP & - & \checkmark & & & & \checkmark & & - & - & \cite{laporte1990selective}\\[1pt] 
PCTSP & - & \checkmark & & & & \checkmark & & - & - & \cite{balas1989prize}\\[1pt]
TSPP & - & \checkmark & & & & & & - & - & \cite{feillet2005traveling}\\[1pt] 
SP & M-M & & & & \checkmark & & & \textit{p\&/d} & \textit{HE} & \cite{anily1992swapping} \\[1pt] 
SPDP & M-M & \checkmark & & & \checkmark & & & \textit{p,d} & \textit{HO} & \cite{ting2013selective} \\[1pt] 
CTSPPD & M-M & & \checkmark & & \checkmark & & & \textit{p,d} & \textit{HO} & \cite{anily1999approximation} \\[1pt]
k-DeliveryTSP & M-M & & \checkmark & & \checkmark & & & \textit{p,d} & \textit{HO} & \cite{chalasani1999approximating} \\[1pt] 
1-PDTSP & M-M & & \checkmark & & \checkmark & & & \textit{p,d} & \textit{HO} & \cite{hernandez2003one} \\[1pt] 
SD1PDTS & M-M & & \checkmark & \checkmark & \checkmark & & & \textit{p,d} & \textit{HO} & \cite{salazar2015split}\\[1pt]
1-TSP-SELPD & M-M & \checkmark & \checkmark & & \checkmark & & & \textit{p,d} & \textit{HO} & \cite{falcon2010one} \\[1pt] 
TSPB & 1-M-1 & & \checkmark & & & & & \textit{p,d} & \textit{HE} & \cite{anily1994traveling} \\[1pt] 
SVRPPD & 1-M-1 & & & \checkmark & \checkmark & & & \textit{p\&/d} & \textit{HE} & \cite{gribkovskaia2007general} \\[1pt] 
SVRPDSP & 1-M-1 & \checkmark & \checkmark & \checkmark & \checkmark & & & \textit{p\&d,d} & \textit{HO} & \cite{gribkovskaia2008single} \\[1pt] 
TSPPD & 1-M-1 & & \checkmark &  & \checkmark & & &  \textit{p\&/d} & \textit{HO} & \cite{mosheiov1994travelling}\\[1pt] 
TSPPD & 1-1 & & & & & & & \textit{p,d} & \textit{HE} & \cite{dumitrescu2010traveling}\\[1pt] 
TSPPDF & 1-1 & & & & & & & \textit{p,d} & \textit{HE} & \cite{erdougan2009pickup}\\[1pt]
TSPPDL & 1-1 & & & & & & & \textit{p,d} & \textit{HE} & \cite{cordeau2010branchL} \\[1pt]
S-DARP & 1-1 & & & & \checkmark & & \checkmark & \textit{p,d} & \textit{HE} & \cite{psaraftis1980dynamic} \\[1pt] 
S-PDPTW & 1-1 & & & & \checkmark & & \checkmark & \textit{p,d} & \textit{HE} & \cite{van1993variable} \\[1pt]
PDTSP & 1-1 & & & & \checkmark & & & \textit{p,d} & \textit{HE} & \cite{kalantari1985algorithm} \\[1pt]
PDTSPH & 1-1 & & & & & & & \textit{p,d} & \textit{HE} & \cite{veenstra2017pickup} \\[1pt]
PDTSPN & 1-1 & & & & & & & \textit{p,d} & \textit{HE} & \cite{gao2023exact} \\[1pt]
m-PDTSP & 1-1 & & & & \checkmark & & & \textit{p\&/d} & \textit{HO} & \cite{hernandez2014multi} \\[1pt]
m1-PDTSP & 1-1 & & & & \checkmark & & & \textit{p\&/d} & \textit{HE} & \cite{hernandez2009multi} \\[1pt]
\midrule
m1-PDSTSP & 1-1 & \checkmark & & & \checkmark & \checkmark & & \textit{p,d} & \textit{HE} & This article \\[1pt] 
\specialrule{1.5pt}{.1em}{.1em} 
\end{tabular}}
\label{tab:comparison_SVRP}
\end{table}


The proposed \textit{m1-PDSTSP} is a 1-1 pickup-and-delivery variant with
heterogeneous requests and selectivity under capacity and route-length
constraints, and each node requires either a pickup or a delivery.
Tab.~\ref{tab:comparison_SVRP} situates \textit{m1-PDSTSP} relative to prior
PDP/TSP variants along these axes.

Algorithmically, these attributes reshape both the search space and the
feasibility geometry: selectivity couples subset choice with routing; 1-1
structure imposes paired, path-dependent constraints; heterogeneous requests
break exchangeability; and tight capacity and route-length restriction create
near-infeasible faces (i.e., boundary regions of the feasible set) where naïve local edits (e.g., a simple swap) fail.
Consequently, effective solution approaches must be feasibility-aware (e.g., paired
destroy-repair, masking) and budget-conscious to operate in real time. 

\subsection{Solution Approaches}\label{sec:SolutionApproaches}

Heuristic algorithms trade off optimality guarantees for speed and scalability
when solving a combination optimization problem (COP). They employ problem-specific hard-coded
rules or general search strategies to quickly construct high-quality solutions
when exact or provably-approximate methods are impractical due to large search
spaces or tight time budgets. The aim is to deliver a feasible, effective
solution within a reasonable timeframe, even if the result is suboptimal.
 
Generally, heuristics are functionally categorized into constructive and
improvement heuristics \citep{kassoumeh2025effect}. We refer to them as
constructors and improvers, respectively. Constructors start with an empty
solution and iteratively build a complete one by adding elements according to a
specific construction rule (often a greedy strategy) selecting the best available
element at each step. While efficient and fast, such approaches may be
trapped in local optima due to their myopic nature. By contrast, improvers begin
with an existing solution and iteratively refine it by exploring its
neighborhood. By systematically modifying the current
solution, they can escape poor initial configurations and often achieve
higher-quality results for a complex combinatorial optimization problems, but
with more computational effort compared with constructors. 

A large body of research on improvers focuses on effective neighborhood
exploration by employing mechanisms such as acceptance ratios in Simulated
Annealing (SA), tabu lists in Tabu Search, and nosing and randomization
strategies in Adaptive Large Neighborhood Search (ALNS) to achieve proper
diversification \citep{pisinger2018large}. While a few studies quantify the
effect of the initial solution
\citep[e.g.,][]{liu2024parallel,kassoumeh2025effect}, most empirical studies
detail operators and stopping conditions, yet keep the initializer fixed or omit
it entirely, leaving its impact underexplored, especially under tight
computational budgets where search trajectories are short. 
Typically, improvers obtain their
initial solutions from either random generation or greedy constructors. Both
methods can rapidly produce complete, feasible solutions from scratch, but they
do not consistently provide high-quality solutions across diverse instances due to
their greedy nature and lack of global perspective. This limitation motivates
learning-based constructors that leverage learned representations of the problem
structure to rapidly generate plausible starts for improvers.


\subsubsection{Learning-based Heuristics}
Neural Combinatorial Optimization (NCO)
solvers or data-driven optimization leverage machine learning techniques to solve COPs. These methods
typically involve training DNNs to learn stochastic policies (an end-to-end
pretrained model) that can generate approximate solutions efficiently. The crux
of learning-based heuristics is their capability to generalize from training
data, allowing them to adapt to new problem instances without the need for
extensive manual tuning. Following \citet{wu2024neural}, NCO methods are
classified into Learning-to-Construct (L2C), Learning-to-Improve (L2I), and
Learning-to-Predict (L2P).

L2C methods \citep[e.g.,][]{vinyals2015pointer,kool2018attention}, including the
SOTA Policy Optimization with Multiple Optima \citep{kwon2020pomo} with diverse
rollouts and data augmentation, build solutions sequentially by selecting the
next unvisited node to extend a partial tour. L2I methods
\citep[e.g.,][]{chen2019learning,ma2021learning} mimic iterative
ruin-and-repair to refine incumbents. L2P approaches
\citep[e.g.,][]{xin2021neurolkh,sun2023difusco} predict
routing-related features such as structure signals or heatmaps that
guide classical optimization algorithms during search. In practice, L2C
typically offers the fastest inference but may require long training and can
struggle to escape local optima; L2I explores richer neighborhoods but can stall
in vast, complex search spaces; and L2P may generalize poorly when instance
sizes or constraint regimes shift \citep{wu2024neural}. Moreover, L2I/L2P are
commonly trained with imitation/supervised signals distilled from high-quality
heuristics or exact solvers, making near-optimal labels costly (or infeasible at
scale) for new problem variants. This challenge is amplified in selective
routing, where the solver must jointly decide which customers to serve and how
to route them under operational constraints. By contrast, L2C methods naturally
accommodate selectivity: (i) they are commonly trained with reinforcement
learning (RL), avoiding reliance on teacher labels; (ii) they can be shaped to
construct profit-maximizing routes under constraints by masking infeasible
actions; and (iii) they can provide more robust cross-regime generalization under
moderate shifts in instance size and constraint tightness by learning global
ordering/selection priors (e.g., long-range pickup-delivery compatibility)
rather than teacher-specific labels. Accordingly, we adopt L2C as the backbone
of our hybrid pipeline.

Existing neural-based methods predominantly address unselective routing problems,
particularly specializing on well-studied instances such as the TSP and vehicle
routing problem (VRP). Conversely, effective NCO methods for PDPs and selective
problems remain under-explored. For PDPs, \cite{li2021heterogeneous} introduce a
heterogeneous attention model (heter-AM) trained with REINFORCE and a greedy
rollout baseline proposed by \cite{kool2018attention}, effectively capturing
precedence and outperforming simple heuristics and neural baselines on PDTSP;
\cite{ma2022efficient} propose Neural Neighborhood Search (N2S) with a
synthesis-attention decoder and tailored remove-reinsert operators to handle
precedence, achieving competitive results versus LKH3 on canonical PDP variants;
\cite{kong2024efficient} develop Neural Collaborative Search (NCS), a hybrid
that couples neural construction and improvement via curriculum and imitation
learning under a shared-critic scheme, outperforming standalone neural methods.
Heter-AM is a L2C approach, but its added complexity yields little benefit
over the attention model of \cite{kool2018attention}. The N2S and NCS methods
are L2I approaches, relying on a heavy Transformer architecture and costly
training. No prior work integrates feasibility masking with a Transformer under a selective PDP setting and none utilizes DL-generated solutions as informed seeds to guide improvement heuristics.

\subsubsection{Large Neighborhood Search}
LNS, introduced by \cite{shaw1998using}, is a powerful improvement metaheuristic
that iteratively destroys and repairs parts of a solution to escape local
optima. Starting from an initial solution, each iteration selects one destroy
and one repair operator: the destroy phase randomly removes a small subset of
customers, and the repair phase reinserts them to improve the objective under
feasibility constraints. Originally developed for the VRP with time windows, LNS
has been adapted to the PDP with time windows (PDPTW) \citep{bent2006two}.
Building on this idea, \cite{ropke2006adaptive} propose ALNS, which maintains
multiple destroy/repair operators and adaptively selects among them based on
observed performance, achieving promising results on PDPTW variants
\citep[e.g.,][]{masson2013adaptive,ghilas2016adaptive}. Operator scores are
updated over time, biasing the selection toward higher-performing pairs as the
search progresses. Unlike standard ALNS/LNS, which destroys requests uniformly, randomly, or by ad hoc scores, our MSLNS uses frequency signals from multiple neural trajectories to identify backbone requests and bias removal accordingly. Coupled with an increasing destroy-size schedule, this creates a learning-driven diversification–intensification mechanism that standard LNS frameworks do not provide. Moreover, prior LNS work largely targets visit-all VRP/PDP settings, whereas selectivity fundamentally changes basin geometry and neighborhood structure, making our integration of neural guidance and selective routing genuinely new.






\section{Problem Formulation}\label{sec:ProbDef}

We formulate the OFEX combinatorial bundling problem as a \textit{m1-PDSTSP}. Each request has exactly one pickup
vertex and one delivery vertex, a demand, and a
revenue. A capacitied vehicle departs from an origin and finishes at a
destination within a route-length limit. The vehicle may serve only a subset of
requests satisfying pickup-before-delivery precedence
and capacity feasibility along the route. Equivalently, this is a capacitated
selective TSP with precedence and paired-demand constraints. The problem is
NP-hard by reduction from Hamiltonian path problem. We now present the formal model (the Mixed Integer Linear Programming formulation is in the Appendix~\ref{appendix:MILP}).



\subsection{Mathematical Formulation}
Let $\mathcal{G}=(\mathcal{V},\mathcal{E})$ be a complete undirected graph,
where $\mathcal{V}=\mathcal{P}\cup\mathcal{D}\cup\{0,2n{+}1\}$ consists of
pickup vertices $\mathcal{P}=\{1,\dots,n\}$, delivery vertices
$\mathcal{D}=\{n{+}1,\dots,2n\}$, a start vertex $0$, and an end vertex
$2n{+}1$. Here $n$ is the selected number of requests from the marketplace. For each request
$h\in\mathcal{H}:=\{1,\dots,n\}$, pickup $i_h\in\mathcal{P}$ is paired with
delivery $i_h{+}n\in\mathcal{D}$ (pickup-before-delivery precedence). Each
request has demand $q_h \in \mathbb{N}$ and revenue $r_h \in \mathbb{R}_{\geq
0}$, collected only if the request is delivered. For each edge
$(i,j)\in\mathcal{E}$, the travel cost (distance) $c_{ij}$ satisfies the
triangle inequality ($c_{ij}\le c_{ik}+c_{kj}$ for all $i,j,k$); edges represent
shortest distances between locations.

\begin{figure}
\centering
\begin{subfigure}{0.19\textwidth}
    \includegraphics[width=\textwidth]{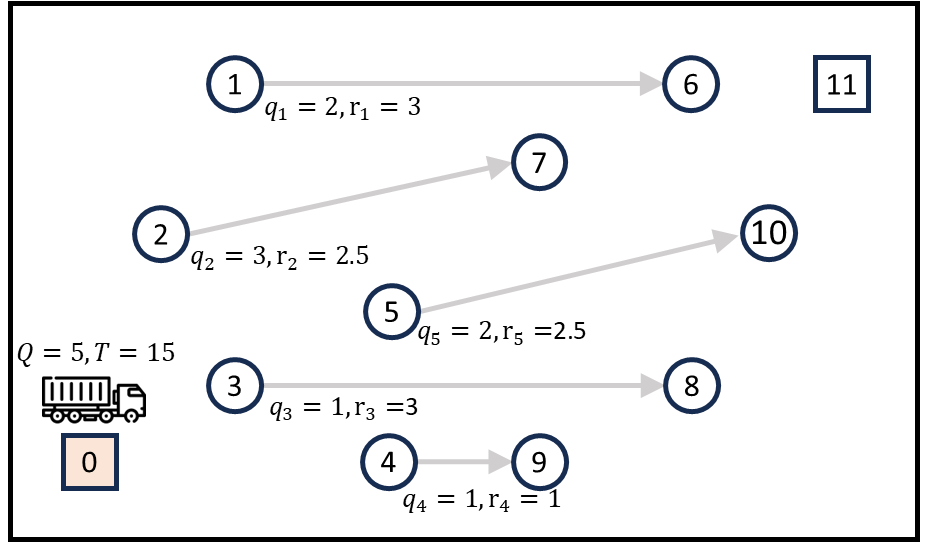}
\end{subfigure}
\begin{subfigure}{0.19\textwidth}
    \includegraphics[width=\textwidth]{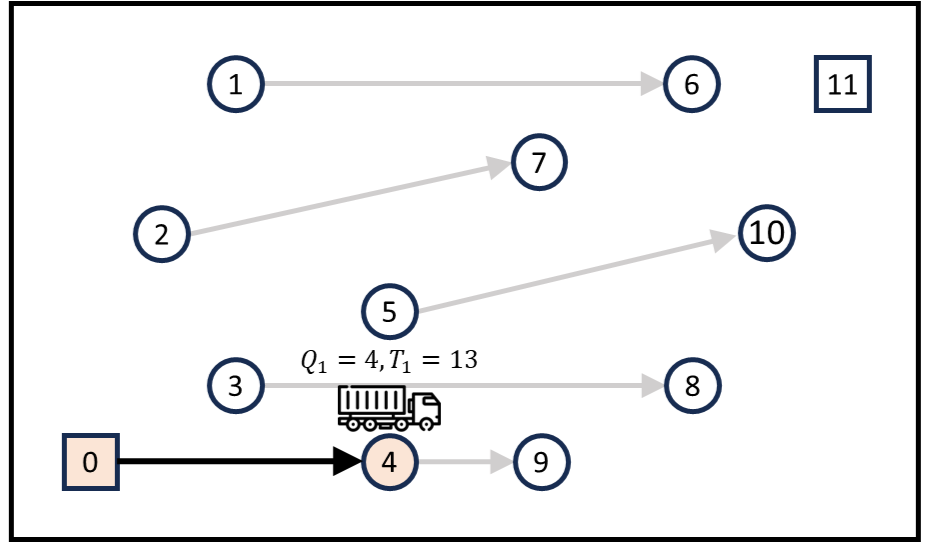}
\end{subfigure}
\begin{subfigure}{0.19\textwidth}
    \includegraphics[width=\textwidth]{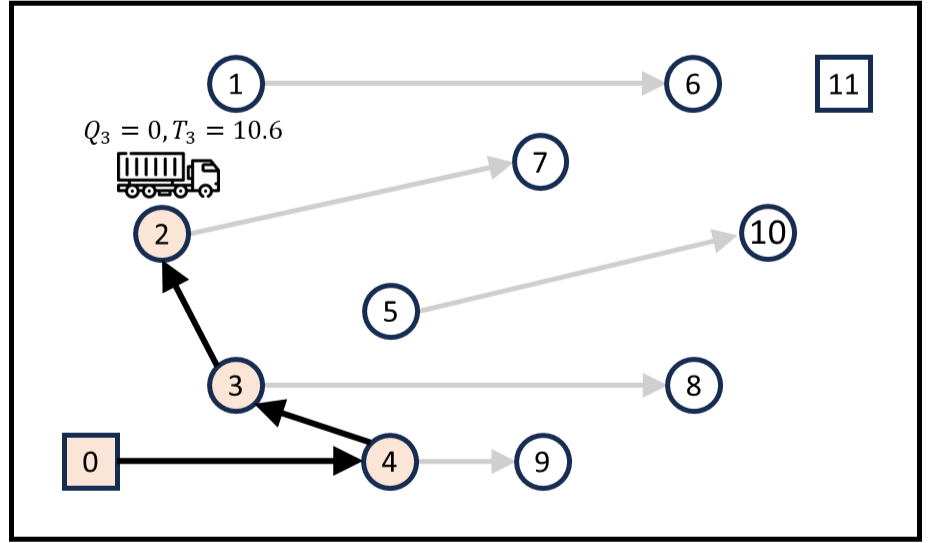}
\end{subfigure}
\begin{subfigure}{0.19\textwidth}
    \includegraphics[width=\textwidth]{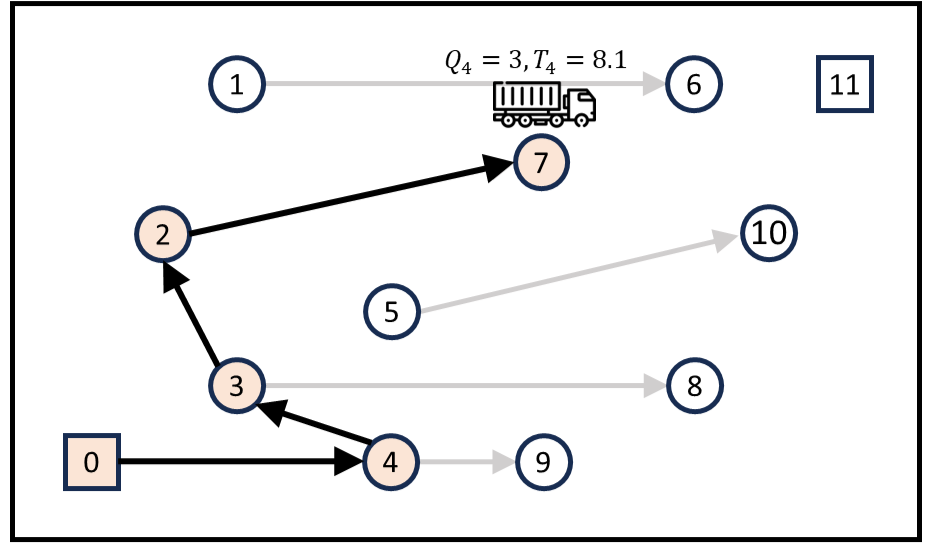}
\end{subfigure}
\begin{subfigure}{0.19\textwidth}
    \includegraphics[width=\textwidth]{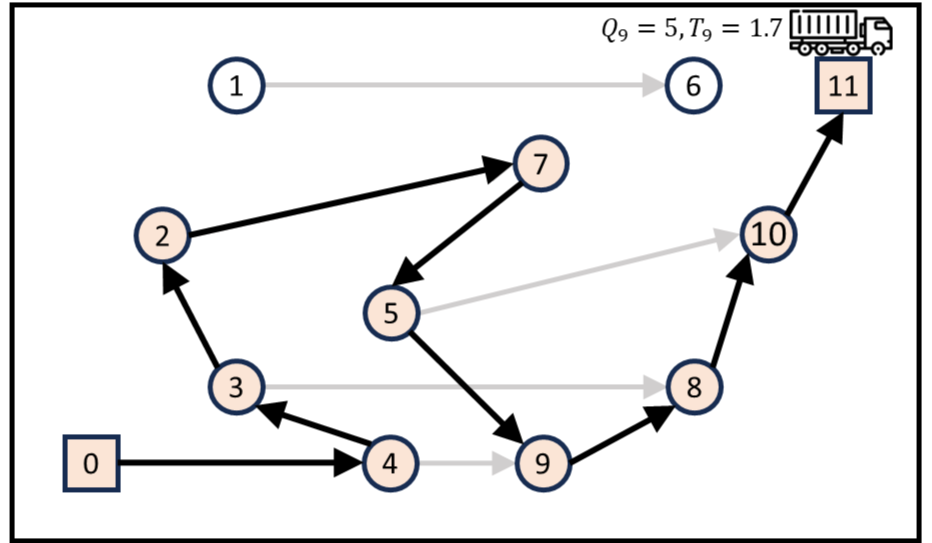}
\end{subfigure}
\caption{Example of the \textit{m1-PDSTSP} with $n=5$ requests and a feasible solution $\pi=(0,4,3,2,7,5,9,8,10,11)$.}
\label{fig:m1-pdstsp}
\end{figure}

We consider one vehicle with capacity $Q$ and a maximum route-length $T$,
starting at $0$ and ending at $2n{+}1$. 
A feasible solution consists of (i) a subset of served requests
$\Pi\subseteq\mathcal{V}$ and (ii) a permutation $\pi$ of $\Pi$ specifying the
sequence of the vertices on the depot-to-depot route. We define that a route
$\pi = \left\lbrace 0 = v_{1}, v_{2}, \dots, v_{\tau} = 2n+1\right\rbrace$ is
feasible if (i) the vehicle stops at most once in each vertex, (ii) the load
capacity on the vehicle never exceeds the vehicle's maximum capacity $Q$, (iii)
the total length of the route does not exceed the prespecified bound ${T}$:
${c(\pi) \leq T}$, where ${c(\pi) = \sum_{i=1}^{\tau-1} c_{v_i,v_{i+1}}}$, and
(iv) if a pickup vertex is visited, then its delivery vertex must be visited
afterwards (not necessarily immediately after). An example is provided in
Fig.~\ref{fig:m1-pdstsp} with $n=5$ requests, $Q=5$, $T=15$ and a feasible
route $\pi=(0,4,3,2,7,5,9,8,10,11)$ with final route-length $13.3$ and collected
revenue $9$. 

\section{Solution Methodology}\label{sec:Methodology} 

To effectively address the \textit{m1-PDSTSP}, we propose a learning-seeded
hybrid search pipeline in Fig.~\ref{alg:mslns} that combines a deep learning
constructor with powerful improvement heuristics. This pipeline has two phases:
the constructor generates high-quality initial solutions rapidly; after
which the improver refines these solutions to approach local optima. This
synergy leverages the strengths of both components, yielding superior
performance on complex routing problems.

In this section, we first elaborate the conceptual motivation behind this design
(Sec.~\ref{sec:motivation}). We then formulate the algorithmic procedure of the
constructor as a MDP (Sec.~\ref{sec:mdp}) and detail its DNN
architecture with an approximate feasibility masking scheme
(Sec.~\ref{sec:AttentionModel}), and training (POMO; Sec.~\ref{sec:training}).
Finally, we describe the improvement heuristic, namely the MSLNS (Sec.~\ref{sec:mslns}).

\subsection{Conceptual Motivation} \label{sec:motivation} 
\begin{figure}
\centering
\begin{subfigure}{0.4\textwidth}
    \includegraphics[width=\textwidth]{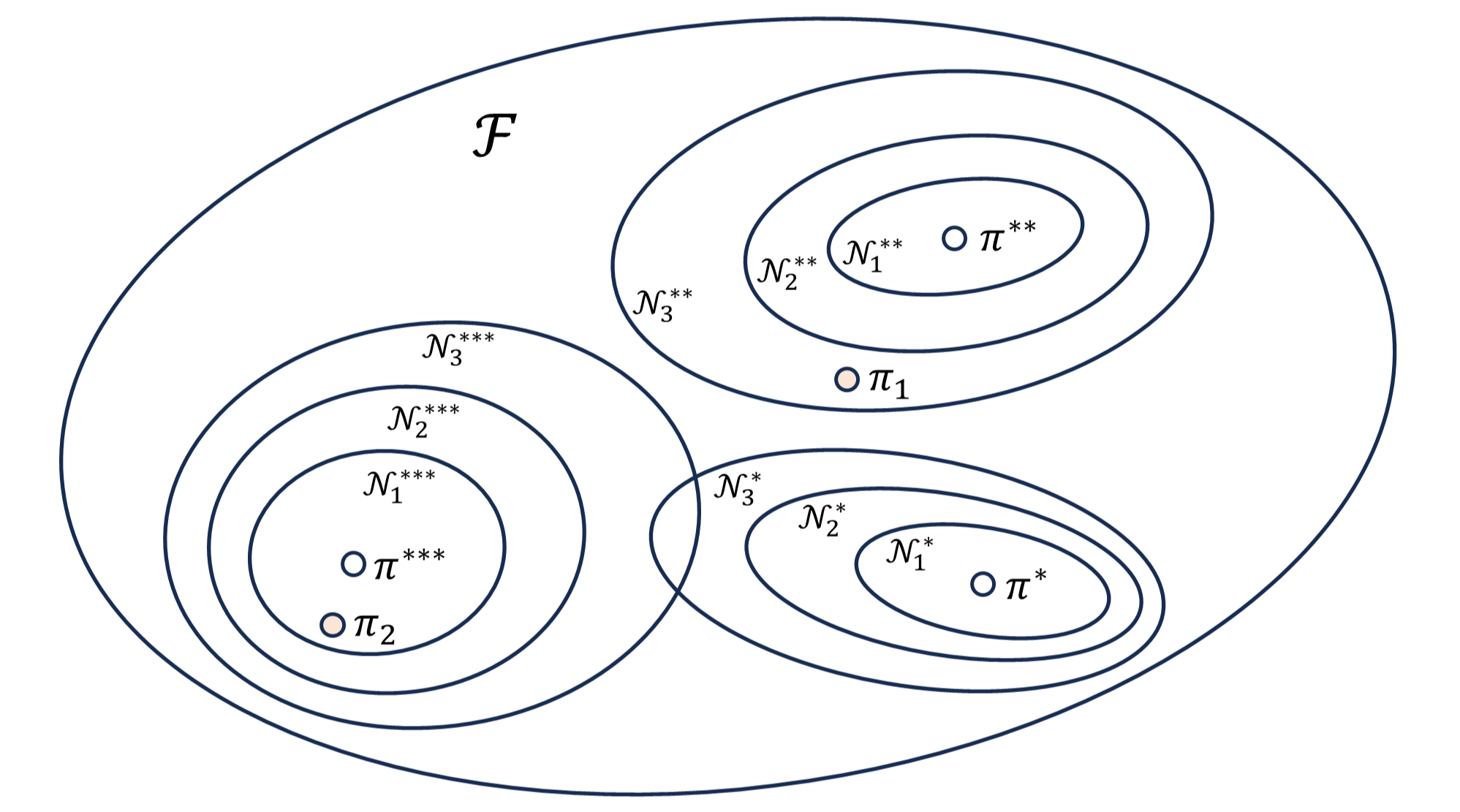}
    \caption{k-attraction basins of local optima}
    \label{fig:LNS_local_optima}
\end{subfigure}
\begin{subfigure}{0.59\textwidth}
    \includegraphics[width=\textwidth]{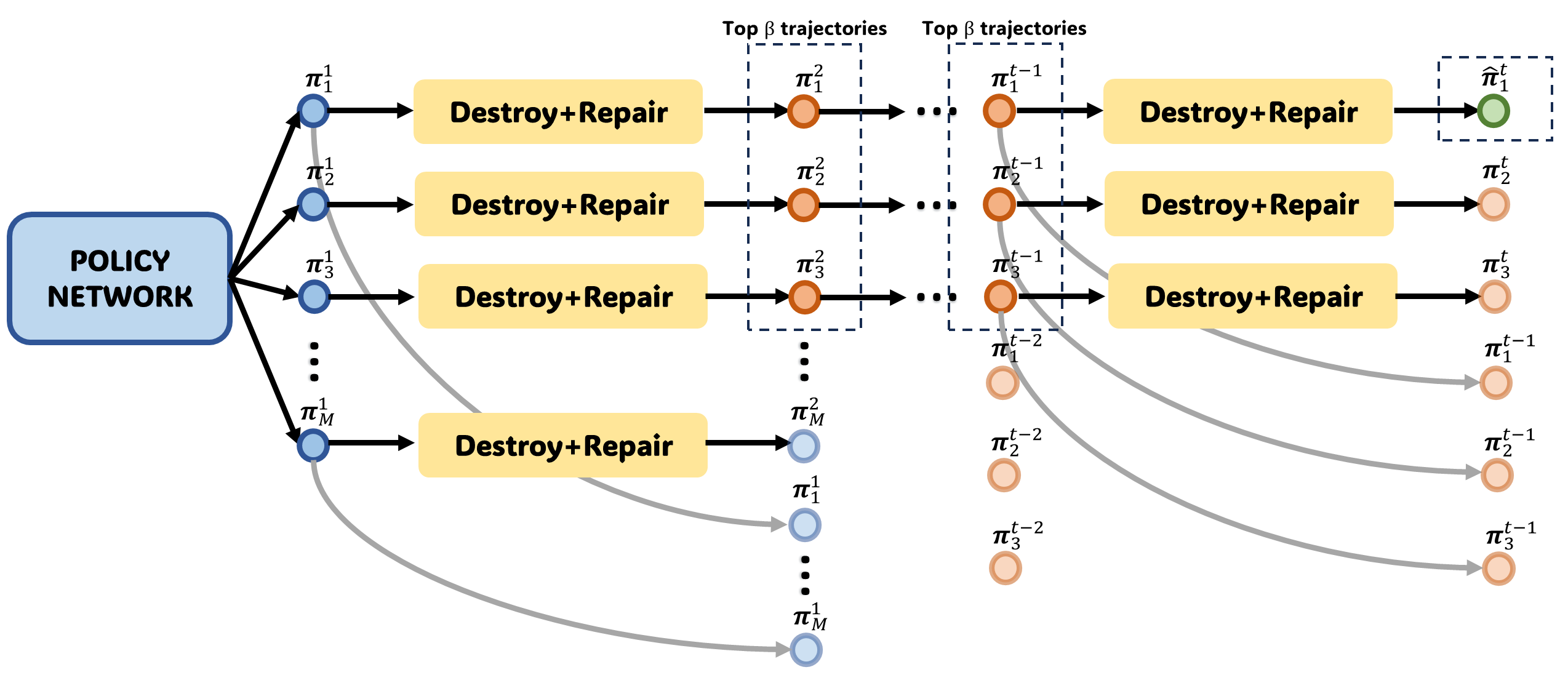}
    \caption{Neural Constructor + MSLNS Improver}
    \label{fig:mslns}
\end{subfigure}
\caption{In the left figure, $\pi^*, \pi^{**}$ are local optima, $\pi^{***}$ is the global optimum ($f(\pi^{*})\leq f(\pi^{**}) \leq f(\pi^{***})$) and $\mathcal{N}_k$ are their related $k$-attraction basins. From initial solutions $\pi_1$ and $\pi_2$, BI-LNS readily reach $\pi^{***}$ from $\pi_2$, whereas $\pi_1$ lies far from the high-quality basins and struggles to reach even $\pi^{**}$. In the right figure, the policy network (Transformer) generates multiple initial solutions in parallel and iteratively refines them with MSLNS to obtain the final solutions (keeping a fixed-width $\beta$ candidates).}
\end{figure}

Given an instance, a local optimal solution is an \textit{attractor} if it is a fixed point of the local search operator (i.e., greedy insertion repair Alg.~\ref{alg:repair} in Appendix~\ref{appendix:baselines}). An \textit{attraction basin} is the subset of the search space from which the local search dynamics converge to the same attractor \citep{jerebic2021novel}. We define $\mathcal{N}_k(\pi)$ as the set of solutions in the $k$-neighborhood of a solution $\pi$, i.e., solutions that can be reached from $\pi$ by applying a destroy-repair operator that removes $k$ requests and reinserts them. Within $\mathcal{N}_k(\pi)$, for any $\pi$ and $k\geq 1$, multiple attractors may exist. A solution $\pi^*$ is called a \textit{$k$-attractor} (with respect to $\mathcal{N}_k$) if $f(\pi^*)\geq f(\pi)$ for all $\pi\in\mathcal{N}_k(\pi^*)$; the corresponding $\mathcal{N}_k(\pi^*)$ is referred to as the \textit{$k$-attractor basin}.
Because the \textit{m1-PDSTSP} is selective, attraction basins are determined by the served-request set. 
Reaching an \textit{attractor} is relatively easy, whereas reaching a superior
\textit{$k$-attractor} is substantially more challenging. The Best-improvement LNS (BI-LNS) improver (Appendix~\ref{appendix:bilns}) with $k$-destroy operator is guaranteed to converge to the \textit{$k$-attractor} as it exhaustively explores the associated \textit{$k$-attractor basin} of the initial solution. If the initial solution lies near a high-quality basin (e.g., $\pi_2$ in Fig.~\ref{fig:LNS_local_optima}), convergence to the global optimum $\pi^{***}$ (where $f(\pi^{***})=\max_{\pi}f(\pi)$) can be achieved with relatively small $k$. Conversely, starting from a distant solution (e.g., $\pi_1$) significantly hampers success even for SOTA metaheuristics, particularly for LNS variants that explore only a portion of the basin. Starting from multiple diverse and high-quality initial solutions can mitigate this issue by probing multiple high-quality basins in parallel. 

Although the solutions produced by the DNN are not guaranteed to be globally optimal or even attractors themselves (see Sec.~\ref{sec:Experiments}), they tend to fall within the attraction region of high-quality basins more often than those generated by classical constructors. Consequently, they provide high-quality initializations that guide downstream improvers with small $k$ toward superior local optima (see Fig.~\ref{fig:attraction_basin_22} in Appendix~\ref{appendix:additional_experiments}), explaining the consistent effectiveness of our hybrid pipeline.

\subsection{Markov Decision Process for Route Construction} \label{sec:mdp} As the constructors construct the solution auto-regressively, the generation process is inherently sequential \citep{drakulic2023bq}. This mirrors the structure of a sequential decision-making problem, where decisions influence both future feasibility and attainable reward. Modeling the decoding procedure as a MDP makes this dependence explicit and provides a principled framework for defining (partial routes), actions (selectable requests), transition rules (updating feasibility), and rewards (incremental revenue).
This formulation clarifies what the neural policy is learning, why masking matters, and where the main computational challenges arise under selective PDP constraints. 

Formally, we denote a COP instance by \(\min_{x \in \mathcal{F}} f(x),\)    
where $\mathcal{F}$ is the (finite) set of feasible, complete solutions, and
$f:\mathcal{F}\to\mathbb{R}$ is the objective (e.g., negative collected
revenue). Equivalently, feasibility can be encoded by a global constraint
predicate $C(\cdot)$ with $\mathcal{F}=\{x: C(x)=\mathrm{T}\}$.

Given a solution space $(\mathcal{X}, \circ, \epsilon, \mathcal{Z})$, where
$\mathcal{X}$ denotes the set of all possible and feasible partial solutions for
a given COP, $\mathcal{Z}$ denotes the set of elementary construction steps
(adding one pickup vertex, one delivery vertex or the end vertex $z^e$), $\circ$
is the concatenation operator and $\epsilon$ is the neutral element (formal definition is in Appendix~\ref{appendix:solution_space}), we consider the set
$\mathcal{S}_\mathcal{X}$ of instances $(f,\mathcal{F})$. The MDP resulting denoted by
$\mathcal{M}_{(f,\mathcal{F})}$ is defined as follows: 

\paragraph{State space:}
\(
\bar{\mathcal{X}}
:= \{ x\in \mathcal{X} : \exists y\in \mathcal{Z}^{\ast}\ \text{with}\ x\circ y \circ z^e\in \mathcal{F} \},
\)
i.e., partial solutions that can be extended (by a finite sequence of steps from $\mathcal{Z}$)
to some feasible complete solution in $\mathcal{F}$.

\paragraph{Action space:}
\(
\mathcal{A}
:= \{ z\in \mathcal{Z} : C(x\circ z)=\mathrm{T}\}
 \cup
\{\epsilon\}
\), which is state dependent, i.e., an action is either a step (adding an unvisited vertex) or the neutral action $\epsilon$.

\paragraph{Deterministic transition:}
 The transition function is deterministic, and can be represented as a labelled
 transition system with the following two rules, with actions serving as transition labels: 
    \begin{equation}
        x \xrightarrow[f(x)-f(x\circ z)]{z} x \circ z \text{ if } x\circ z \in \bar{\mathcal{X}} \;\ , \ \; x\xrightarrow[0]{\epsilon} x  \text{ if } x \in \mathcal{F}.
    \label{eq:mdp_rules}
    \end{equation}
    Here $x\in \bar{\mathcal{X}}$ is a state and $z \in \mathcal{Z}$ is a step. 

\paragraph{Reward function:} $r(x,z) = f(x)-f(x')$ where $x,x'\in
\bar{\mathcal{X}}$, and $x' = x \circ z $ for all $z \in \mathcal{Z}$ or $x' =x
$ when $z=\epsilon$. Starting from $x_0=\epsilon$, an episode applies admissible
actions until reaching a terminal state $x_\tau\in\mathcal{F}$. Because we
\emph{minimize} $f$, maximizing the cumulative return $\sum_{t=1}^{\tau}
r(x_t,a_t)= f(x_0)-f(x_\tau)$ is equivalent to driving $f(x_\tau)$ as low as
possible.


To minimize the objective function $f$ of the \textit{m1-PDSTSP} defined in (\ref{eq:obj2}), 
the stochastic policy $p_{\boldsymbol{\theta}}$ automatically selects a vertex
at each time step based on a learned distribution under the operational
constraints. This process is repeated iteratively until the policy chooses to
visit the end vertex. The policy produces a sequence of the selected vertices
that specifies the vehicle's visit order, $\pi = (z_1 \circ z_2 \circ \cdots
\circ z_{\tau}  )$, where $z_1$ and $z_\tau$ are the start and end vertices of
the vehicle, respectively. Based on the chain rule, the probability of an output
solution is factorized as follows,
\begin{equation}
    p_{\boldsymbol{\theta}} (\pi \,|\, s) = \prod_{t=1}^{\tau} p_{\boldsymbol{\theta}}(z_t\,|\,\boldsymbol{z}_{1:t-1},s), \quad \forall s \in \mathcal{S}_{\mathcal{X}}. 
\end{equation}


\subsubsection{MDP-Transformer Integration}
\begin{figure}
\centering
\begin{subfigure}{0.32\textwidth}
    \includegraphics[width=\textwidth]{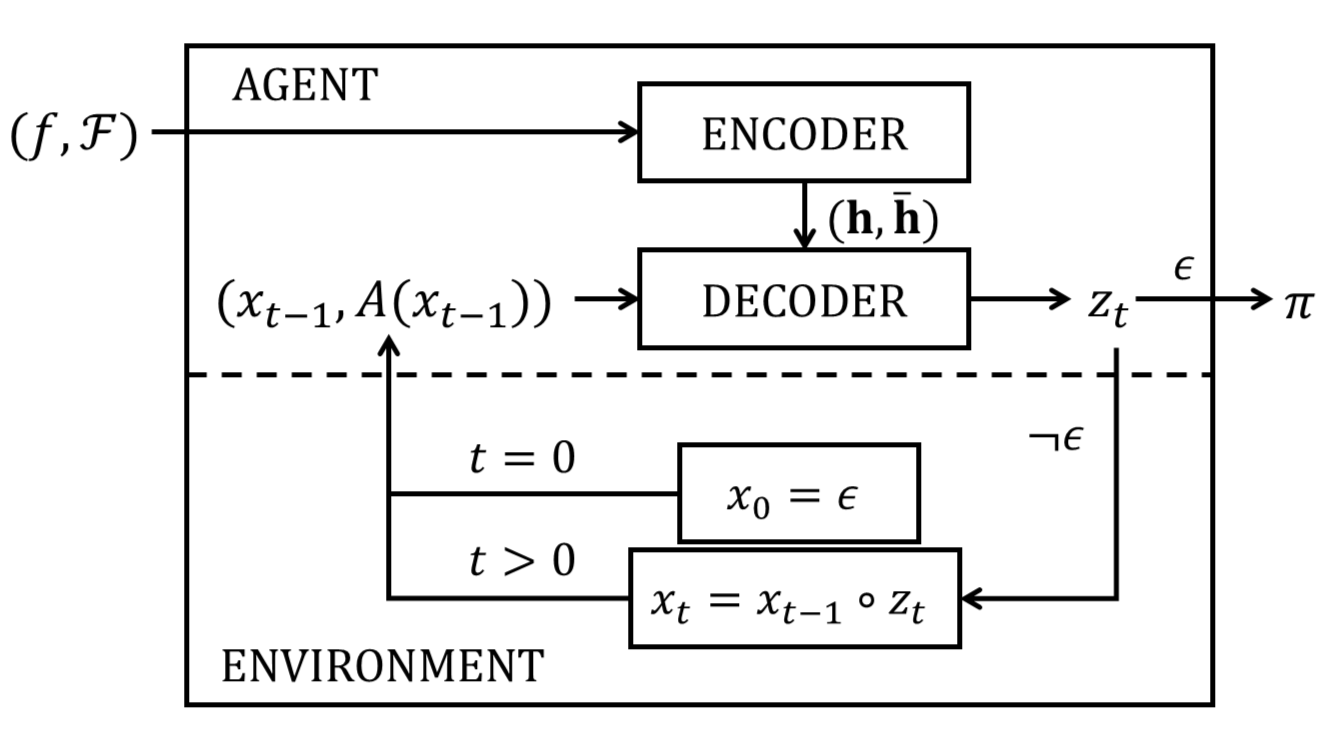}
    \caption{MDP-Transformer}
\end{subfigure}
\begin{subfigure}{0.32\textwidth}
    \includegraphics[width=\textwidth]{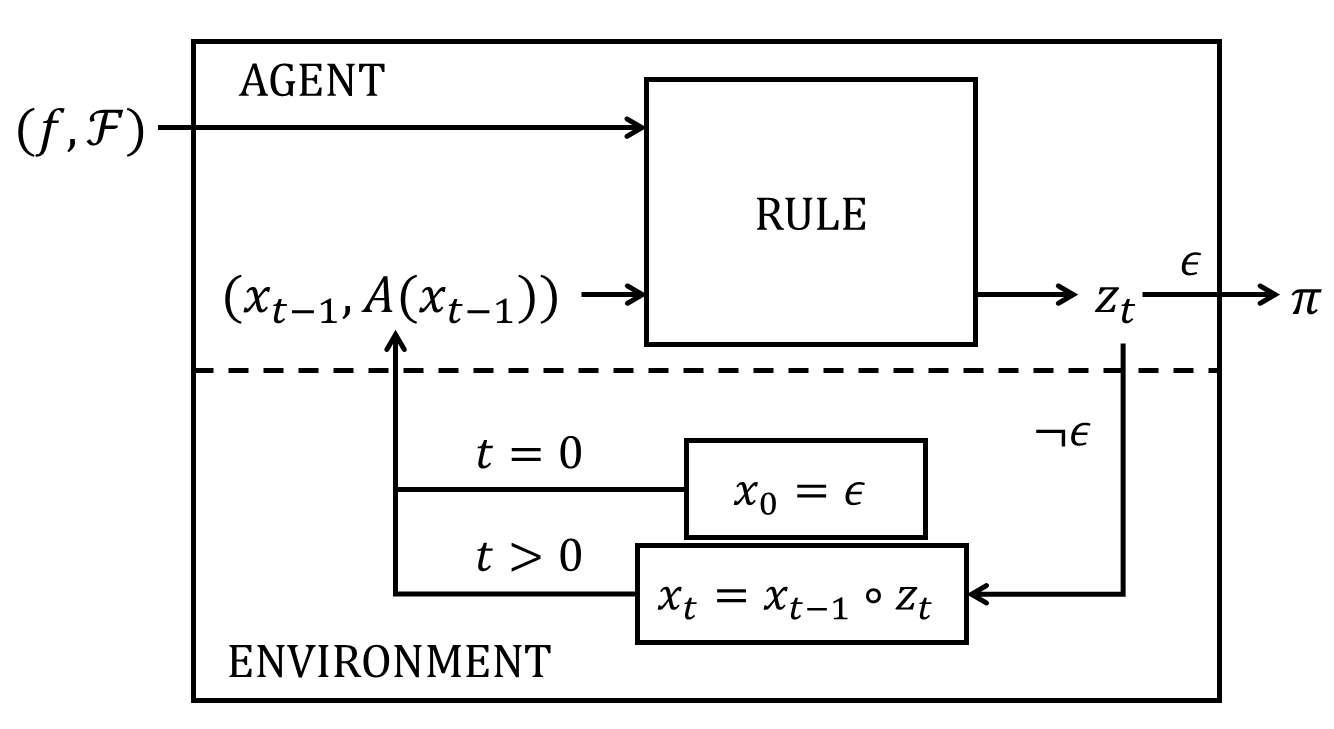}
    \caption{MDP-Constructor}
\end{subfigure}
\begin{subfigure}{0.32\textwidth}
    \includegraphics[width=\textwidth]{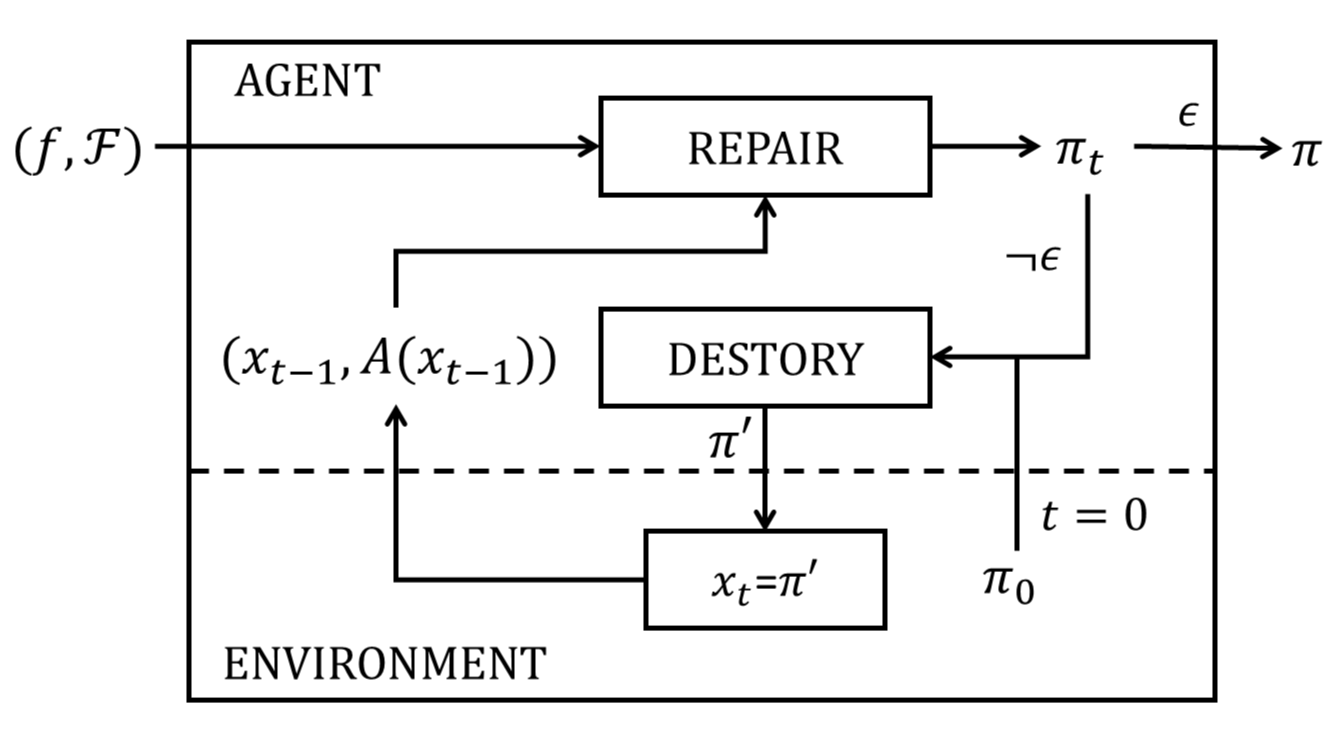}
    \caption{MDP-Improver}
\end{subfigure}
\caption{Illustrations of MDP integrating with a Transformer, a constructor and an improver (Appendix~\ref{appendix:mdp_improvers}).}
\label{fig:mdp}
\end{figure}

In Fig.~\ref{fig:mdp}, we couple the MDP with the
Transformer by formalizing the state at time $t$ as
\(
\tilde{x}_t \;=\; \bigl(\mathbf{h},\, \bar{\mathbf{h}},\, x_t\bigr),
\)
where $\mathbf{h}$ and $\bar{\mathbf{h}}$ are the node embeddings and the global
instance embedding learned by the encoder (which are shared across all time
steps, see details in Sec.~\ref{sec:encoder}), and $x_t =
\boldsymbol{z}_{1:t-1}$ is the partial solution maintained up to time $t$. The
action $z_t$, generated by the decoder, is the next feasible vertex to visit.
The reward \( r_t = f(x_{t-1}) - f(x_t). \) captures the one-step improvement in
the objective. An episode terminates when the policy selects the end vertex,
i.e., $z_\tau = z^e$. The total return is thus
\(
R = \sum_{t=1}^{\tau} r_t = f(\epsilon) - f(\boldsymbol{z}_{1:\tau}) = -f(\pi).
\)
As for the constuctive metaheuristics, the state
at time $t$ is $x_t$ and the policy is rule-based (i.e., fitness). 

\subsubsection{Computational Challenges}
In the constructive process, evaluating the action space at each step requires
validating the feasibility of adding each unvisited vertex to the current partial
solution, formally, for each $z \in \mathcal{Z}$, validating whether $C(x\circ
z)=\mathrm{T}$. This is computationally expensive because even validating a
single capacity-feasible pickup involves verifying that its paired delivery, all
yet-undelivered vertices, and the end vertex can still be visited within
route-length limits, an NP-hard, shortest-Hamiltonian-path problem (see
Fig.~\ref{fig:example_routes}, fourth panel; requests 3 and 4 remain
undelivered, and validating whether picking up request 1 or 3 is admissible
requires such a test). Because this validation may be performed for
every candidate action at every construction step, it effectively embeds an
NP-hard subproblem inside the inner loop, causing the computational cost to
scale combinatorially with the number of undelivered vertices and unvisited
pickup vertices. Repeatedly solving such validations during the training of the
neural networks is intractable without aggressive approximation or pruning. To
accelerate this, we propose an approximate feasibility masking scheme (in
Sec.~\ref{sec:decoder}) that screens candidates based on cheap lower bound
surrogates. This validation also happens in repair operator, which motivates an efficient searching algorithm.

\subsection{Details of Attention Model}\label{sec:AttentionModel} To enable the hybrid pipeline, we require a neural constructor capable of producing strong, feasibility-aware initial solutions. Accordingly, our attention model
builds on \citet{kool2018attention} and \citet{kwon2020pomo}; we customize both the encoder-decoder Transformer architecture and
the policy-gradient training procedure for the \textit{m1-PDSTSP}.
\subsubsection{Encoder} \label{sec:encoder}
The encoder maps instance features into contextualized embeddings via
self-attention, producing keys/values for the decoder to score feasible actions.

\paragraph{Node feature embeddings:}
From the $d_{\mathbf{x}}$-dimensional input features $\mathbf{x}_i$, the
encoder computes initial $d_{\mathbf{h}}$-dimensional node embeddings
$\mathbf{h}_i^{(0)}$ ($d_{\mathbf{h}} = 128$) via a learned
type-specific linear map:
\begin{equation}
    \scalebox{0.9}{$
    \mathbf{h}_i^{(0)} = \begin{cases}
        W^{\mathbf{x}}_\text{depot} \Big[ \mathbf{y}_i, T \Big] + \mathbf{b}_\text{depot}^{\mathbf{x}} &i=0,2n+1\\
        W^{\mathbf{x}}_\text{pickup} \Big[\mathbf{y}_i, \mathbf{y}_{i+n}, \hat{q}_i, \hat{r}_i \Big] +  \mathbf{b}_\text{pickup}^{\mathbf{x}} &i=1, \dots, n\\
        W^{\mathbf{x}}_\text{delivery} \Big[\mathbf{y}_i, -\hat{q}_i, \hat{r}_i \Big] +  \mathbf{b}_\text{delivery}^{\mathbf{x}} &i=n+1, \dots, 2n,\\
    \end{cases}
    $}
\end{equation}
The input features include the node
coordinates $\mathbf{y}_i\in \mathbb{R}^2$, normalized demand $\hat{q}_i \in [0,1]$ if $i$ is a pickup
node and $-\hat{q}_i$ if $i$ is a delivery node, and normalized revenue
$\hat{r}_i \in [0,1]$ (normalization ensures that all inputs share a comparable scale).

Afterwards, the node embeddings go through $L$
attention layers, each of which encompasses a multi-head attention (MHA)
sublayer ($M=8$ heads with dimensionality $d_{\mathbf{h}}/ M = 16$) and a feed-forward (FF) sublayer (one hidden layer with dimension $512$ and ReLu
activation). We denote with
$\mathbf{h}_i^{(\ell)}$ the node embeddings produced by layer $\ell \in \{1,
\dots, L\}$. Similar to \cite{kool2018attention}, each sublayer adds a
skip-connection and batch normalization (BN):
\begin{equation}
    \scalebox{0.9}{$
    \hat{\mathbf{h}}_i = \textbf{BN}^\ell \Big(\mathbf{h}_i^{(\ell-1)} + \textbf{MHA}_i^\ell\Big(\mathbf{h}_0^{(\ell-1)}, \dots, \mathbf{h}_{2n+1}^{(\ell-1)}\Big)\Big), \quad
    \mathbf{h}_i^{(\ell)} = \textbf{BN}\Big(\hat{\mathbf{h}}_i + \textbf{FF}^\ell(\hat{\mathbf{h}}_i)\Big) .
    $}
\end{equation}

\paragraph{Graph embedding:} The encoder computes an aggregated embedding $\bar{\mathbf{h}}^{(L)}$
of the input instance as the mean of the node embeddings at the final layer
$\bar{\mathbf{h}}^{(L)} = \frac{1}{2n+2}\sum_{i=1}^{2n} {\mathbf{h}}^{(L)}_i$.

\subsubsection{Decoder} \label{sec:decoder}
The decoding is auto-regressive and happens sequentially. At timestep $t \in
\{1,\dots 2n+1\}$, the decoder generates a probability for selecting a node
$z_t$ based on the embeddings ($\mathbf{h}_i^{(\ell)}, \mathbf{h}_i^{(\ell)}$) from the
encoder and the outputs $\boldsymbol{z}_{1:t-1}$ generated before time $t$.

\paragraph{Operational Constraints:} To facilitate the capacity and maximum
length constraints, we keep track of the remaining normalized capacity $\hat{Q}_{t}$ and
remaining length $T_t$ at time $t$. At $t=1$, $\hat{Q}_1 = 1$ and $T_1 =
T$, thereafter, they are updated as follows:
\begin{equation}
    \scalebox{0.9}{
    $
    \hat{Q}_{t+1} = \max(\hat{Q}_t - \hat{q}_{z_t}, 0)$, \quad $
    T_{t+1} = T_t - c_{z_{t-1}, z_t},
    $
    }
\end{equation}
where $ c_{z_{t-1}, z_t}$ is the distance from node $z_{t-1}$ to $z_t$ and we define $z_{0} = 0$.

\paragraph{Context Embedding:} The context for the decoder for
\textit{m1-PDSTSP} at time $t$ includes the node embeddings of the current location $z_{t-1}$ and the end
location of the vehicle $2n+1$, the remaining capacity $\hat{Q}_t$, and the
remaining route length ${T}_t$: 
$
    \mathbf{h}_{(c)}^{(L)} = [\bar{\mathbf{h}}^{(L)}, \mathbf{h}^{(L)}_{z_{t-1}}, \mathbf{h}^{(L)}_{2n+1}, \hat{Q}_t, {T}_t],
$
where $[\cdot, \cdot]$ is the horizontal concatenation operator. We compute a
new context node embedding $\mathbf{h}_{(c)}^{(L+1)}$ using the MHA mechanism
(described in Appendix~\ref{appendix:attention_model}), where the context
embedding serves as the query. The keys and values are derived from the node
embedding $\mathbf{h}^{(L)}_i$, allowing the context to aggregate relevant
information from all nodes through attention (\( \boldsymbol{q}_{(c)}^{(L)} =
W^Q \mathbf{h}_{(c)}^{(L)},\, \boldsymbol{k}_i = W^K \mathbf{h}_i^{(L)},\,
\boldsymbol{v}_i = W^V \mathbf{h}_i^{(L)} \)).

\paragraph{Masking:} During decoding, all invalid nodes are dynamically masked
at each step to ensure solution feasibility. Upon departing from the start node,
all delivery nodes are initially masked. If there are undelivered deliveries,
the end node is masked. After visiting a pickup node, its corresponding delivery
node is unmasked, while all visited nodes (incl. the start node) are masked.
Pickup nodes are also masked if their demand exceeds the current remaining
capacity. Finally, each unvisited pickup node $j$ is masked if serving $j$,
completing all yet-undelivered deliveries, and reaching the end node would
violate the remaining route-length limit. In principle, validating this last
condition requires, at each step and for each unvisited pickup node, solving a
shortest Hamiltonian path problem, an NP-hard problem that is prohibitively
expensive to perform millions of times during training. We therefore design a
lower-bound screening: for each pickup node $j$, we compute a cheap lower bound
surrogate on the route length using only the current node $j$, its delivery node
$j+n$, and the end depot $2n+1$; if this surrogate already exceeds the remaining
time limit, $j$ is masked. This avoids the hard feasibility validation and
reduces the per-candidate test from $\mathcal{O}(n^2)$ by 2-Opt (which provides an approximation of the minimum route length) to $\mathcal{O}(1)$. The trade-off is that some infeasible pickup nodes may not be masked and be selected
by the policy. 

We denote the resulting potential cost of visiting a pickup
node $j$ at step $t$ by $\tilde{c}_j$ as follows:
\begin{equation}
    \scalebox{0.9}{$
    \tilde{c}_j = \begin{cases}
        c_{z_{t-1},j}+c_{j,j+n}+c_{j+n,2n+1} & \text{if training,}\\
        c_{z_{t-1},j}+\text{TwoOpt}_{j,2n+1}(S)& \text{if inference},
    \end{cases}
    $}
\end{equation}
where $\text{TwoOpt}_{s,t}(S)$ returns the tour obtained by applying
2\text{-}Opt to a path constrained to start at $s$ and end at $t$ while visiting
all nodes in $S$; here $S \subseteq \{n+1, \dots, 2n\}$ denotes all
yet-undelivered delivery nodes including $j+n$. Initialized from a
nearest-neighbor tour, 2\text{-}Opt typically removes detours and substantially
shortens the route. We apply it as an more accurate validator, executed in parallel
across the batch. 
Given that the lower bound may be loose, some infeasible nodes
may remain unmasked, we shape the reward (objective) function to penalize
infeasible solutions:
\begin{equation}
    \scalebox{0.9}{$
    \tilde{f}(\pi) = f(\pi) + \rho \cdot \max\{0, c(\pi) - T\},
    $}
\end{equation}
where $\rho$ is a penalty coefficient and $c(\pi)$ is the
total distance/length of the route $\pi$.

We compute the compatibility of the query with all nodes based on
(\ref{eq:compatibility}) in Appendix~\ref{appendix:attention_model}, and mask
nodes which are not feasible. We define the mask matrix
$\mathcal{M}_t$ at time $t$ for all $j \in V$ as follows:
\begin{equation}
    \scalebox{0.9}{$
     \mathcal{M}_t[j] = \begin{cases}
         -\infty  & \text{if } \exists t' < t: \pi_{t'} = j \;\text{or}\; \not\exists t' < t: \pi_{t'} = \max\{0,j-n\} 
         \;\text{or}\; \hat{q}_{j,t} >\hat{Q}_t \;\text{or}\; \tilde{c}_j > {T}_t, \\
         1 & \text{otherwise}.
     \end{cases} 
     $}
\end{equation}

Finally, we add one final single attention head decoder layer to compute the
probability vector $p_{\boldsymbol{\theta}}$. The next node is sampled from the
$p_{\boldsymbol{\theta}}$ using the categorical distribution during training:
\begin{equation}
    \scalebox{0.9}{$
    p_i = p_{\boldsymbol{\theta}}(z_t = i \,|\, s, \boldsymbol{z}_{1:{t-1}}) = \text{softmax}\bigg(C \cdot \tanh \Big(\frac{{\boldsymbol{q}_{(c)}^{(L+1)}}^\top \boldsymbol{k}_j}{\sqrt{d_k}} \odot \mathcal{M}_t \Big)\bigg),
    $}
\end{equation}
where $C=10$, $d_k = \frac{d_{\mathbf{h}}}{M}$ is the query and key dimensionality, and $\odot$ denotes the Hadamard (element-wise) product. This process iteratively continues untill the end node is visited.

\subsubsection{Training Algorithm} \label{sec:training} During training, the
problem instances are independently and identically distributed from a
distribution $\mathbb{S}$. We adopt the SOTA training algorithm, Policy
Optimization with Multiple Optima (POMO) \citep{kwon2020pomo}, to train multiple
greedy trajectories. Our motivation differs from the original paper: POMO was
designed for unselective, sequence invariant problems (e.g., TSP), where a
symmetry-aware policy should produce the same solution regardless of the start
node, and its optimizer exploits this symmetry. In our selective setting, solutions are not sequence invariant. The starting
action significantly affects feasibility and quality, as a poor start can
accumulate deviations throughout decoding. We therefore leverage POMO's
multi-start strategy to emphasize start-node selection and broaden exploration
over the solution space. 

POMO begins with the vehicle start node and selects $M$ different pickup nodes
$\{z_1^1, z_1^2, \dots, z_1^M\}$ to start for exploration, where $z_1^j \in \mathcal{P}$,
$j \in \{ 1, \dots, M \}$, and $M \leq |\mathcal{P}|$. The $M$ start pickup
nodes are the top $M$ choices of the stochastic policy $p_{\boldsymbol{\theta}}$ given by the current
policy network. As our problem is selective and anchored at a fixed start node,
the full POMO multi-start over all nodes is unnecessary. Accordingly, we
restrict POMO to $M=n/2$ parallel starting positions (one per pickup) rather
than $|V|$. Then the network samples $M$ solution trajectories $\{\pi^1, \pi^2,
\dots, \pi^M \}$ via Monte-Carlo simulation by drawing $B$ i.i.d. sample instances
$s_1, \dots, s_B\sim\mathbb{S}$. Based on (\ref{eq:monte-carlo sampling
reinforce}) in Appendix~\ref{appendix:training_algorithm}, in POMO, we apply the
shared baseline (to reduce variance and stabilize training) for all $j \in M$ and $i \in B$: 
\begin{equation}
    b^j(s_i) = b_{\text{shared}}(s_i) = \frac{1}{M} \sum_{k=1}^{M} \tilde{f}(\pi^k_i). 
\end{equation}
The training procedure is summarized in Alg.~\ref{alg:pomo}. 

\begin{algorithm}
\caption{Policy Optimization with Multiple Optima (POMO) Training}
\footnotesize
\begin{algorithmic}[1]
\State \textbf{Input}: number of epochs $E$, steps per epoch $K$, batch size $B$, number of start nodes $M$
\State Initialize policy parameters $\boldsymbol{\theta}$
\For{epoch $= 1,\dots,E$} \Comment{outer training loop}
  \For{step $= 1,\dots,K$} \Comment{mini-batch updates per epoch}
    \State $s_i \leftarrow \textsc{RandomInstance}() \ \ \forall i\in\{1,\dots,B\}$ \Comment{sample a batch of instances}
    \State $\{\pi^1_{1,i}, \pi^2_{1,i}, \dots, \pi^M_{1,i} \} \leftarrow \textsc{SelectStartPickupNodes}(s_i) \quad \forall i \in \{1, \dots, B\}$ \Comment{$M$ diverse starts per instance}
    \State $\pi^j_i \leftarrow \textsc{SampleRollout}(\pi_{1,i}^j,s_i, p_{\boldsymbol{\theta}}) \quad \forall i \in \{1, \dots, B\}, \ \forall j \in \{1,\dots,M\}$ \Comment{roll out a full solution from each start}
    \State $b_i \leftarrow \frac{1}{M} \sum_{j=1}^{M} \tilde{f}(\pi_i^j) \quad \forall i \in \{1,\dots, B\} $ \Comment{per-instance baseline: mean objective over $M$ starts}
    \State $\nabla \mathcal{L} \leftarrow \frac{1}{BM}\sum_{i=1}^{B}\sum_{j=1}^{M}\big(\tilde{f}(\pi_i^j)-b_i\big)\nabla_{\boldsymbol{\theta}}\log p_{\boldsymbol{\theta}}(\pi_i^j)$ \Comment{REINFORCE gradient with baseline}
    \State $\boldsymbol{\theta} \leftarrow \textsc{Adam}(\boldsymbol{\theta}, \nabla \mathcal{L})$ \Comment{optimizer update}
  \EndFor
\EndFor
\end{algorithmic}
\label{alg:pomo}
\end{algorithm}

\subsection{Multi-Start Large Neighborhood Search} \label{sec:mslns}


To apply a multi-start strategy, a straightforward way is to run LNS independently on each trajectory; however, this is computationally inefficient and fails to exploit the structural information provided by the DNN.
Because the neural constructor generates multiple greedy trajectories independently, the frequency with which a request appears across these trajectories reflects its relative importance (or lack of substitutability) within the instance. To leverage this information, we design a softmax-biased removal strategy that selects requests to destroy simultaneously across all trajectories. We prioritize high-frequency requests for early removal because they typically repair back into the solution, revealing backbone structure that can be fixed so that subsequent search concentrates on escaping low-quality basins rather than rediscovering stable route components. Moreover, the destroy size controls the search regime: small destroy sizes favor intensification within a basin, whereas larger sizes enable transitions across basins.

For illustration, consider 10 requests with a global optimum $\{0,3,5,7,9\}$ and a local $1$-attractor $\{0,1,2,3\}$. Destroying $1$ first leads repair back to the same attractor, when each request is destroyed at most once (as in our experiments to ensure efficient exploration), the subsequent search can never reach the global optimum. By contrast, destroying $0$ first allows transitions (e.g., via subsequent removals of $\{1,2\}$, $\{2,3\}$, or $\{1,3\}$) that open paths to basins containing the global optimum.

Based on these insights, we design MSLNS with the following components: 
(i) destroy operator with softmax-biased removals and a short-term memory of
deleted requests to avoid redundant deletions; (ii) an increasing destroy-size
schedule; (iii) greedy insertion repair operator; and (iv) multi-start
diversification.
Specifically, the destroy step first deduplicates identical
served-request sets across the candidates, then selects requests that have not
yet been marked for destruction by sampling from a softmax over their
frequencies across all trajectories (temperature $\alpha=1$; larger $\alpha$
biases more toward frequent requests). We gradually increase the destroy size
from $2$ up to $k_{max}=4$. The repair step selects best-insertion greedily
(Alg.~\ref{alg:repair} in Appendix~\ref{appendix:baselines}). Because running
LNS on all trajectories is costly, we adopt a pruning strategy that retains a
fixed-width set of top $\beta$ candidates at each iteration, and we further
apply a fast 2-Opt pass to improve the travel distance for each candidate. The
overall procedure appears in Fig.~\ref{fig:mslns} and Alg.~\ref{alg:mslns}. In
practice, MSLNS only destroys a request at most once because of the time budget
and the high-quality initial solutions produced by the policy network. We choose
POMO ($M$ parallel greedy rollouts as Alg.~\ref{alg:pomo-inference} in
Appendix~\ref{appendix:baselines}) as the constructor due to its effectiveness
in training and its natural ability to produce multiple high-quality, diverse
greedy trajectories for MSLNS. Through extensive experiments in next section, we
verify the robustness of MSLNS with different constructors and the effectiveness
of each component in it.

\begin{algorithm}[t]
\caption{Multi-Start Large Neighborhood Search (MSLNS)}
\footnotesize
\begin{algorithmic}[1]
\Require $M$ initial trajectories $\Pi$, beam width $\beta$, reward function $R$, time limit $t_{\max}$, temperature $\alpha$, max removals $k_{\max}$
\State $\mathcal{U}\gets\varnothing$; \quad $i\gets1$ \Comment{Memory of removed requests}
\While{time $<t_{\max}$}  \Comment{Main search loop until time limit}
  \State $(\mathbf{u},\mathbf{c})\gets \textsc{CountUnique}(\Pi)\setminus \mathcal{U}$ \Comment{Find unremoved unique requests and their counts}
  \State \textbf{if} $(\mathbf{u},\mathbf{c})=\varnothing$ \textbf{then break} \ \textbf{end if} \Comment{Stop if no requests left to destroy}
  \State $\mathbf{p}\gets \mathrm{softmax}(\alpha\,\mathbf{c})$;\quad $\mathbf{u}^\star\sim \mathrm{Multinomial}(\mathbf{p},\min\{i+1,k_{\max}\})$ \Comment{Sample $k$ requests to destroy based on weighted counts}
  \State $\mathcal{U}\gets \mathcal{U}\cup\mathbf{u}^\star$ \Comment{Record destroyed requests in memory}
  \State $\Pi'\gets \textsc{Repair}(\textsc{Destroy}(\Pi,\mathbf{u}^\star))$ \Comment{Apply destroy-repair to generate new candidate trajectories}
  \State $\Pi'\gets \textsc{TwoOpt}(\textsc{Dedup}(\Pi'))$  \Comment{Local improvement and duplicate removal}
  \State $\Pi\gets \textsc{Top}_\beta(\Pi\cup \Pi',R)$\Comment{Keep top-$\beta$ solutions by reward}
  \State $i\gets i+1$ \Comment{Increase removal budget for next iteration}    
\EndWhile
\State \Return $\arg\max_{\pi\in \Pi} R(\pi)$
\end{algorithmic}
\label{alg:mslns}
\end{algorithm}

\section{Experiments}\label{sec:Experiments} 
In this section, we investigate how initialization quality, attraction basin structure, and neighborhood design jointly determine search effectiveness in selective PDPs under tight latency constraints.
Our empirical study has three goals: (i) characterize the strengths and limitations of classical heuristics, neural constructors, and hybrid methods; (ii) quantify the contribution of each MSLNS component; and (iii) assess robustness across instance scales and problem regimes. To this end, we evaluate our POMO+MSLNS approach on the \textit{m1-PDSTSP} against the following neural baselines: AM \citep{kool2018attention}, POMO \citep{kwon2020pomo}, and
POMO-SGBS \citep{choo2022simulation}; heuristic baselines: GS, Hill Climbing (HC), Multi-Start Greedy Search (MSGS), BI-LNS, LNS, and SA; and Gurobi v12.0.3. All baselines are implemented and tuned for our setting
in Appendix~\ref{appendix:baselines}. Then, we conduct ablations to quantify the contributions of individual MSLNS components. Additional analyses, including sensitivity to revenue parameters, route-length limits, and zero-shot generalization, are reported in Appendix~\ref{appendix:additional_experiments}. These results further demonstrate the robustness of the learning-seeded hybrid search design and the strong cross-regime generalization of POMO+MSLNS.

\subsection{Experimental Setup}
We detail instance generation, training protocol, and
inference/time-budget settings.

\paragraph{Instance Generation:} All instances are randomly generated.
Following the settings in \cite{li2021heterogeneous}, the location of the depots
and customer (pickup and delivery pair) nodes are randomly and independently
drawn from a uniform distribution in the unit square $[0,1]^2$. The
traveling cost between two nodes are measured by Euclidean distance. The demand
of each customer node is uniformly sampled from the set $\{2,3,4,5\}$. For the
distribution of the revenue, we consider four variants: \textit{Distance},
\textit{Ton-Distance}, \textit{Uniform} and \textit{Constant}. We normalize the
revenue $r_h$ such that $\hat{r}_h \in [0,1]$ for all $h \in \mathcal{H}$. For
the \textit{Distance} setting, the revenue is proportional to the distance
between the pickup and delivery nodes. This distance-based normalization mirrors
distance-driven earnings/costs and keeps results comparable across instances
(the details of latter three are in
Appendix~\ref{appendix:additional_experiments}). We sample capacity $Q\sim
\text{Unif}\{8,\dots,20\}$ and set the route-length budget \( T \sim \text{Unif}
\{ \max\{2,\ \lceil \frac{3n}{20} \, c_{0,2n+1} \rceil\},\dots, \max\{2,\ \lceil
\frac{10n}{20} \, c_{0,2n+1} \rceil\} \}\), anchoring $T$ to the start-end depot
distance $c_{0,2n+1}$ and scaling with instance size $n$. This enables $Q$
commensurate with per-stop demands, enabling nontrivial consolidation without
trivial full coverage, while the range of $T$ yields time budgets from tight to
relaxed, ensuring that capacity and route-length constraints frequently
influence feasibility. Regarding the scale, the problem sizes are set to $N =
\{22, 42, 82, 122\}$ (incl. two depots and all nodes of pickup and delivery),
and are denoted by PDSTSP22/42/82/122.


\paragraph{Training:}
To compare AM and POMO fairly, we train both on the instances randomly generated
from the same generator. AM follows \cite{kool2018attention} with $2500$ batches
$\times$ $512$ instances per epoch; POMO employs $2000$ batches $\times$ $64$ instances
per epoch. Both share the same policy network (encoder depth $N=3$) and apply the
Adam with initial learning rate $10^{-4}$ (and weight decay $10^{-6}$ for POMO),
without learning-rate decay. All runs execute on a single NVIDIA A100 GPU (120 GB
memory). To compute them equally, we set the number of training epochs as in
Tab.~\ref{tab:training_hyperparameters}, so that each method receives approximately
the same wall-clock time.
\begin{table}[h!]
\centering
\caption{Training hyperparameters}
\setlength{\tabcolsep}{2.5pt}      
\renewcommand{\arraystretch}{1.05} 
\resizebox{\columnwidth}{!}{
\begin{tabular}{c c c c c}
\hline
    Instance & PDSTSP22 & PDSTSP42 & PDSTSP82 & PDSTSP122 \\
    \hline
    AM & 150 epochs (20 min/epoch) & 125 epochs (40 min/epoch) & 150 epochs (80 min/epoch) & 125 epochs (120 min/epoch) \\
    POMO & 200 epochs (15 min/epoch) & 200 epochs (25 min/epoch) & 200 epochs (60 min/epoch) & 150 epochs (100 min/epoch) \\
    \hline
\end{tabular}}
\label{tab:training_hyperparameters}
\end{table}
\vspace{-0.5\baselineskip}

\paragraph{Inference:} 
For each problem, we report performance on a test set of 1,000 instances
generated with the same procedure as the training set but different random
seeds. All experiments were run on CPU only, using a 32-vCPU machine, and we
report the corresponding runtimes. For most algorithms, runtime and performance
can be traded off by adjusting hyperparameters. Since reporting full trade-off
curves is impractical, we instead select representative configurations that
strike a reasonable balance between runtime and solution quality. The chosen
hyperparameters are summarized in Tab.~\ref{tab:inference_baselines} in
Appendix~\ref{appendix:baselines}. Since larger instances make each repair more
expensive, we scale the per-instance time limit $t_{max}$ with instance size
($=0.5s/1s/4s/10s$ for PDSTSP22/42/82/122). 



\subsection{Experimental Results}\label{sec:m1-pdstsp} Our main experiments
focus on the \emph{Distance} setting, as it has the most practical
significance. We define a metric Winning Rate (WR) to measure the percentage
of instances where an algorithm achieves the best-known solution among all
compared methods (including the exact Gurobi solver when available). WR captures how often
a method attains top performance and also reflects how frequently its solutions fall within the $0$-attraction basin of the best-known solution, complementing the optimality gap metric. A higher WR indicates more consistent excellence across instances. We
can only attain the optimum for small instances using Gurobi within practical runtime.
\begin{table*}[t]
\centering
\caption{Performance matrix: PDSTSP22/42/82/122 on \textit{Distance}. }
\label{tab:main_results_distance}
\small
\setlength{\tabcolsep}{2pt}
\renewcommand{\arraystretch}{0.95}
\resizebox{\textwidth}{!}{%
\begin{tabular*}{1.5\textwidth}{@{\extracolsep{\fill}}
    c l
    r r r r r
    r r r r r   
    r r r r r   
    r r r r r   
@{}}
\toprule
& & \multicolumn{5}{c}{\textbf{PDSTSP22}} & \multicolumn{5}{c}{\textbf{PDSTSP42}} & \multicolumn{5}{c}{\textbf{PDSTSP82}} & \multicolumn{5}{c}{\textbf{PDSTSP122}} \\
\cmidrule(lr){3-7}\cmidrule(lr){8-12}\cmidrule(lr){13-17}\cmidrule(lr){18-22}
 & \multicolumn{1}{c}{\textbf{Algorithm}} &
\multicolumn{1}{c}{Tim} & \multicolumn{1}{c}{Rev} & \multicolumn{1}{c}{Pro} & \multicolumn{1}{c}{Gap} & \multicolumn{1}{c}{WR} &
\multicolumn{1}{c}{Tim} & \multicolumn{1}{c}{Rev} & \multicolumn{1}{c}{Pro} & \multicolumn{1}{c}{Gap} & \multicolumn{1}{c}{WR} &
\multicolumn{1}{c}{Tim} & \multicolumn{1}{c}{Rev} & \multicolumn{1}{c}{Pro} & \multicolumn{1}{c}{Gap} & \multicolumn{1}{c}{WR} &
\multicolumn{1}{c}{Tim} & \multicolumn{1}{c}{Rev} & \multicolumn{1}{c}{Pro} & \multicolumn{1}{c}{Gap} & \multicolumn{1}{c}{WR} \\
\midrule
\multirow{8}{*}{\rotatebox{90}{\small Heuristics}}
&GS+2-Opt                      & 10s & 1.80 & -0.55 & 18.76\% & 17.8\% & 2m  & 3.83 & 0.08 & 24.55\% & 3.0\%  & 12m  & 7.93 & 1.48 & 28.79\% & 0.3\%  & 50m  & 12.59 & 2.84 & 30.17\% & 0.0\% \\
&GS+HC                         & 17s & 1.92 & -0.53 & 13.19\% & 28.6\% & 3m  & 4.18 & 0.19 & 17.50\% & 6.8\%  & 13m  & 8.66 & 1.64 & 22.19\% & 0.9\%  & 53m  & 13.63 & 2.94 & 24.42\% & 0.0\% \\
&GS+BNS$^{t_{max}}$             & 8m  & 2.04 & -0.46 & 8.74\% & 41.9\% & 12m & 4.45 & 0.43 & 12.24\% & 11.6\% & 1.4h & 9.14 & 2.10 & 17.86\% & 2.1\%  & 3.6h & 14.42 & 3.73 & 19.99\% & 0.6\% \\
&GS+LNS$^{t_{max}}$            & 82s & 2.02 & -0.47 & 7.91\% & 44.6\% & 18m & 4.42 & 0.40 & 12.86\% & 11.8\% & 1.5h & 9.09 & 2.05 & 18.31\% & 2.1\%  & 3.5h & 14.79 & 4.10 & 17.94\% & 0.7\% \\
&GS+SA$^{t_{max}}$             & 9m  & 2.14 & -0.37 & 3.49\% & 70.0\% & 23m & 4.38 & 0.37 & 13.65\% & 15.6\% & 1.6h & 8.96 & 1.93 & 19.49\% & 3.3\%  & 3.8h & 14.18 & 3.50 & 21.31\% & 1.2\%  \\
&GS+SA                         & \multicolumn{5}{c}{--}   & 2.8h & 4.70 & 0.68 & 7.25\% & 32.6\% & 3.3h & 9.50 & 2.46 & 14.66\% & 7.6\%  & 9.4h & 14.88 & 4.19 & 17.43\% & 2.9\% \\
&MSGS                          & 53s & 2.01 & -0.48 & 9.51\% & 36.8\% & 11m & 4.26 & 0.26 & 16.04\% & 7.3\%  & 1.7h & 8.81 & 2.26 & 20.91\% & 1.6\%  & 6.7h & 13.92 & 4.05  & 22.77\% & 0.1\% \\
&MSGS+MNS$_{[5,3]}^{t_{max}}$  & 4m  & 2.17 & -0.32 & 2.20\% & 74.9\% & 29m & 4.73 & 0.73 & 6.80\%  & 27.5\% & 3.1h & 9.68 & 2.66 & 13.05\% & 6.2\%  & 10.5h& 15.25 & 4.58 & 15.40\% & 2.6\% \\
\midrule
\multirow{9}{*}{\rotatebox{90}{\footnotesize NCOs}}
&POMO-single                   & 6s  & 1.85 & -0.57 & 16.76\% & 18.8\% & 49s & 4.48 & 0.56 & 11.67\% & 4.6\%  & 5m   & 10.10 & 3.18 & 9.27\% & 0.8\%  & 11m  & 16.69 & 6.22 & 7.42\% & 0.1\% \\
&AM                            & 7s  & 1.97 & -0.45 & 11.29\% & 28.7\% & 45s & 4.54 & 0.62 & 10.49\% & 7.2\%  & 5m   & 10.13 & 3.21 & 9.03\% & 1.1\%  & 12m  & 16.52 & 6.02 & 8.34\% & 0.2\% \\
&AM-sample                     & 66s & 2.03 & -0.37 & 8.38\% & 35.8\% & 8m  & 4.74 & 0.79 & 6.53\%  & 12.7\% & 1.7h & 10.63 & 3.67 & 4.52\% & 4.0\%  & 5.6h & 17.34 & 6.83 & 3.80\% & 2.1\%\\
&AM-MS                         & 54s & 2.10 & -0.36 & 5.22\% & 47.5\% & 7m  & 4.79 & 0.83 & 5.53\%  & 14.3\% & 1.7h & 10.61 & 3.67 & 4.68\% & 4.9\%  & 5.4h & 17.24 & 6.74 & 4.38\% & 0.7\% \\
&AM-BS                         & 59s & 2.13 & -0.33 & 5.22\% & 55.7\% & 8m  & 4.82 & 0.86 & 4.90\%  & 19.4\% & 1.7h & 10.66 & 3.71 & 4.28\% & 6.0\%  & 6.1h & 17.27 & 6.77 & 4.20\% & 1.6\% \\
&POMO                          & 54s & 2.08 & -0.38 & 6.00\% & 43.8\% & 7m  & 4.79 & 0.83 & 5.50\%  & 15.7\% & 1.7h & 10.67 & 3.72 & 4.18\% & 4.4\%  & 5.4h & 17.44 & 6.95 & 3.26\% & 1.5\% \\
&POMO-SGBS$_{[5,5]}$           & 3m  & 2.09 & -0.40 & 5.88\% & 52.6\% & 40m & 4.83 & 0.83 & 4.77\%  & 23.5\% & 3.0h & 10.82 & 3.84 & 2.77\% & 14.4\% & 9.8h & 17.75 & 7.23 & 1.52\% & 19.5\%\\
&AM-sample$_{[100]}$           & 11m & 2.07 & -0.37 & 6.48\% & 41.4\% & 1.3h& 4.85 & 0.87 & 4.38\%  & 16.4\% & \multicolumn{5}{c}{--} & \multicolumn{5}{c}{--}  \\
&AM-BS$_{[100]}$               & 10m & 2.13 & -0.33 & 3.94\% & 56.5\% & 1.3h& 4.88 & 0.91 & 4.38\%  & 23.9\% & \multicolumn{5}{c}{--} & \multicolumn{5}{c}{--}  \\
\midrule
\multirow{6}{*}{\rotatebox{90}{\footnotesize Hybrids}}
&AM+HC                         & 23s & 2.05 & -0.40 & 7.69\% & 40.2\% & 2m  & 4.70 & 0.73 & 7.38\%  & 14.3\% & 20m  & 10.44 & 3.47 & 6.22\% & 3.8\%  & 1.3h & 16.96 & 6.44 & 5.94\% & 1.4\% \\
&AM+BNS$^{t_{max}}$             & 87s & 2.13 & -0.37 & 4.00\% & 57.0\% & 11m & 4.85 & 0.85 & 4.26\%  & 28.8\% & 1.3h & 10.65 & 3.64 & 4.34\% & 11.8\% & 4.4h & 17.20 & 6.62 & 4.57\% & 8.1\%  \\
&AM-BS+HC                      & 75s & 2.15 & -0.31 & 3.02\% & 62.0\% & 9m  & 4.89 & 0.92 & 3.48\%  & 28.2\% & 1.9h & 10.83 & 3.87 & 2.73\% & 11.9\% & 7.2h & 17.53 & 7.05 & 2.76\% & 5.1\% \\
&AM-BS+BNS$^{t_{max}}$          & 2m  & 2.18 & -0.31 & 1.83\% & 71.6\% & 17m & 4.97 & 0.96 & 2.07\%  & 41.1\% & 2.9h & 10.93 & 3.92 & 1.83\% & \textbf{26.4\%} & 10.3h & 17.67 & 7.11 & 1.96\% & 17.9\% \\
&POMO+HC                       & 71s & 2.12 & -0.34 & 4.26\% & 53.5\% & 10m & 4.90 & 0.93 & 3.38\%  & 27.1\% & 1.9h & 10.86 & 3.90 & 2.43\% & 9.6\%  & 6.6h & 17.74 & 7.27 & 1.61\% & 4.5\% \\
&POMO+BNS$^{t_{max}}$           & 2m  & 2.16 & -0.34 & 2.80\% & 64.6\% & 17m & 4.97 & 0.96 & 2.00\%  & 40.1\% & 2.8h & 10.98 & 3.98 & 1.40\% & 23.0\% & 9.8h & 17.86 & 7.30 & 0.90\% & \textbf{20.7\%} \\
\midrule
\multirow{3}{*}{\rotatebox{90}{\footnotesize Ours}}
&AM-MS+MNS$_{[5,3]}^{t_{max}}$ & 5m  & 2.19 & -0.30 & 1.19\% & 78.9\% & 25m & \textbf{4.99} & \textbf{1.01} & \textbf{1.55\%} & \textbf{44.0\%} & 3.2h & 10.93 & 3.97 & 1.84\% & 19.9\% & 10.0h & 17.73 & 7.24 & 1.62\% & 12.5\% \\
&POMO+MNS$_{[5,3]}^{t_{max}}$  & 4m  & 2.19 & -0.30 & 1.35\% & 78.1\% & 28m & 4.99 & 1.00 & 1.65\%  & 42.3\% & 3.1h & 10.97 & 4.01 & 1.43\% & 18.7\% & 9.8h & 17.86 & 7.38 & 0.91\% & 11.7\% \\
&POMO+MNS$_{[10,5]}^{t_{max}}$ & 7m  & \textbf{2.20} & \textbf{-0.30} & \textbf{0.96\%} & \textbf{82.6\%} & 40m & 4.99 & 1.00 & 1.57\%  & 43.2\% & 3.9h & \textbf{10.99} & \textbf{4.03} & \textbf{1.24\%} & 20.4\% & 12.4h & \textbf{17.88} & \textbf{7.40} & \textbf{0.80\%} & 12.4\% \\
\midrule
\multirow{2}{*}{\rotatebox{90}{\footnotesize BLs}}
&POMO+MNS$_{[10,10]}$          & 15m & 2.20 & -0.29 & 0.57\% & 88.2\% & 3.3h& 5.07 & 1.08 & 0.00\%  & 65.8\% & 12.7h & 11.13 & 4.17 & 0.00\% & 48.5\% & 1.4d & 18.03 & 7.54 & 0.00\% & 38.3\% \\
&Gurobi (1000s)                & 17.1h & 2.22 & -0.31 & 0.00\% & 97.7\% & \multicolumn{5}{c}{--} & \multicolumn{5}{c}{--} & \multicolumn{5}{c}{--}  \\
\bottomrule

\end{tabular*}}

\smallskip
\footnotesize{BL: Baseline. Tim: Time, Rev: Revenue, Pro: Profit, Gap: Gap on Revenue. WR: Winning Rate. MNS: MSLNS, BNS:BI-LNS.\\
Profit (=Revenue-Length) is positively correlated with Revenue. Some LS run finish before the
time budget due to early convergence, while some MSLNS exceed $t_{max}$ when
completing the final repair. Across all four scales,
POMO+MSLNS$_{[10,5]}^{t_{max}}$ generally achieves the best performance—among
time-budgeted methods—but typically incurs higher runtime than other
baselines.}
\end{table*}


Given Tab.~\ref{tab:main_results_distance}, we observe:
(i) among the heuristic baselines, MSG+MSLNS$^{t_{max}}_{[5,3]}$ ($M=5$, $\beta=3$, under $t_{max}$ time limit) attains the best performance on all scales, highlighting the effectiveness of multi-start exploration even without learned guidance;
(ii) neural methods significantly outperform classical heuristics with lower wall-clock time (especially at larger scales), underscoring their superior scalability;
(iii) hybrid methods that combine neural constructors with LNS consistently outperform standalone neural methods (even with sophisticated search), which often fail to reach an \textit{attractor} as even simple HC can further improve them;
(iv) POMO+MSLNS$^{t_{max}}_{[10,5]}$ generally outperform all the other methods ($< 2\% $ gaps on total revenue and high WRs across all scales), demonstrating that high-quality neural initialization and structured multi-start neighborhood exploration are complementary;
(v) POMO+MSLNS$_{[10,10]}$ nearly matches Gurobi on PDSTSP22 (gap=0.57\% and WR=88.2\%), confirming that POMO+MSLNS can approach exact-solver quality;
(vi) POMO variants outperform AM variants at larger scales, suggesting that POMO’s multi-start training yields more robust policies for large-scale selective routing;
(vii) AM consistently achieve higher WR than GS, and neural methods equipped with sophisticated search attain higher WR than neighborhood search initialized from GS, indicating that neural constructors reach the $0$-attraction basin of the best-known solution more frequently than classical constructors, thereby providing more reliable starting points for downstream improvement;
(viii) AM-BS+BI-LNS$^{t_{max}}$ and POMO-SGBS$_{[5,5]}$ achieve high WR at all scales, suggesting that they explore attraction basins that are systematically different from those reached by MSLNS.
While not achieving the lowest gap, they intermittently outperform by discovering instance-specific structures not captured by the latter.

Taken together, these results demonstrate that the hybrid pipelines dominates
the quality--latency Pareto frontier (see Fig.~\ref{fig:model_selection_42} in
Appendix~\ref{appendix:additional_experiments}) on the \textit{m1-PDSTSP}.
Practically, AM/AM+HC are plausible choices under tight budget, whereas
POMO+BI-LNS and POMO+MSLNS should be the default for middle/high budget. The modest gains of
MSLNS$^{t_{max}}$ over BI-LNS$^{t_{max}}$ arise because, for large $T$, MSLNS
is less efficient and often cannot converge within the budget; hence, it
can underperform BI-LNS (consistent with the evidence in Tab.~\ref{tab:gap_results}
in Appendix~\ref{appendix:gap_size}). 

\begin{figure}[h]
\centering
\includegraphics[width=\textwidth]{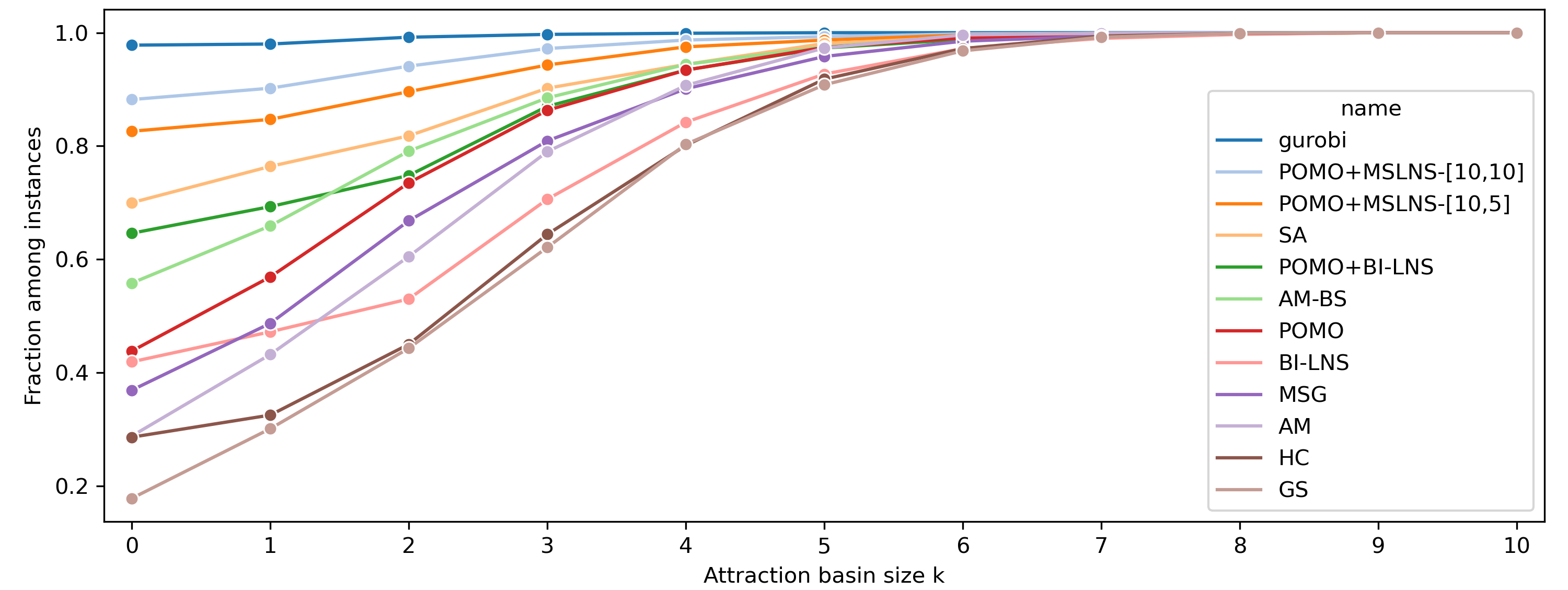}
\caption{Fraction of solutions in the $k$-attraction basin of the best-known solution on PDSTSP22. Higher values at smaller $k$ indicate closer proximity to high-quality attraction basins. Hybrid methods exhibit significantly stronger basin alignment than classical metaheuristics.}
\label{fig:attraction_basin_22}
\end{figure}

Fig.~\ref{fig:attraction_basin_22} demonstrates that POMO+MSLNS consistently attains the highest fraction across all values of $k$ on PDSTSP22, indicating that its solutions most frequently lie within the $k$-attraction basin of the best-known solution. In contrast, classical metaheuristics require substantially larger neighborhoods to reach the same basin. Among constructors, AM consistently outperforms HC and GS, while POMO and AM-BS dominate MSG and BI-LNS across all $k$. Notably, MSG outperforms BI-LNS for small neighborhoods, highlighting the benefit of multi-start strategies. SA performs competitively in this setting due to the small instance size and relatively loose computational budget.

Overall, the results establish the
necessity of the combination of a neural constructor and a heuristic improver to
tackle the challenging \emph{m1-PDSTSP}, highlight the superior
performance of POMO+MSLNS method, and underscore the importance of high-quality initialization and multi-start diversification for efficiently accessing high-quality attraction basins.

\subsection{Ablation Study} \label{sec:ablation_study}
In this section, we examine the importance of MSLNS
components (softmax-biased removal, multi-start strategies, increasing destroy
size) using a small set of 200 instances on \emph{Distance} PDSTSP42. For a fair
comparison, we fix the instance set and employ the same pretrained POMO for all
variants. Each variant is run five independent times with different solver
seeds to smooth out stochastic fluctuations from both the model and the search process. We report the mean. 

\noindent\textbf{MSLNS variants.} (i) \textit{Softmax+Inc-k (Full)}:
softmax-biased removal, multi-start, increasing destroy size ($k_{max}=4$); (ii)
\textit{Uniform+Inc-k}: replace softmax with uniform removal, keep increasing
size ($k_{max}=4$); (iii) \textit{Softmax+Fixed-k}: softmax removal with fixed
$k=4$; (iv) \textit{Uniform+Fixed-k}: uniform removal with fixed $k=4$; (v)
\textit{Single-start}: set $M=1$, others as in \textit{Full}; (vi) 2-Opt+LNS:
apply a fast 2-Opt after the best solution given by POMO, then plain LNS with
uniform removal with random $k\sim\text{Unif}\{1,2,3\}$, no multi-start.
We also report: (i) \textit{BI-LNS-only} and (ii) \textit{2-Opt+BI-LNS}.    



\begin{figure}[h]
\includegraphics[width=\textwidth]{images/experiments/ablation_study.png}
\vspace{-2em}
\caption{Ablation of MSLNS components on PDSTSP42 (\emph{Distance}): mean
revenue across all instances, running over five seeds. Methods marked “(T)” are
unlimited-time variants (run to convergence). The unlimited full MSLNS—Softmax+Inc-$k$ (T)—achieves the best performance.}
\label{fig:ablation_study}
\end{figure}
\vspace{-0.5\baselineskip}

Fig.~\ref{fig:ablation_study} presents that the full MSLNS (Softmax+Inc-$k$)
achieves the best mean performance, both under time limits (mean=$4.710$) and at
convergence (mean=$4.744$).
Both components matter: fixing $k$ or replacing
softmax with uniform removal degrades quality, though limited-time variants
narrow the gap slightly. Multi-start also helps: Single-start
and 2-Opt+LNS
drops notably (mean $<4.700$). We further observe that a 2-Opt pass materially
strengthens the LNS improver; BI-LNS-only (mean=$4.629$) underperforms relative to 2-Opt+BI-LNS
(mean=$4.691$) across budgets. We note that the variance across five seeds is
small, yielding stable behavior and consistent rankings. Overall, top
performance requires the full configuration and dropping any component
consistently lowers quality under both time-limited and converged runs.

\section{Conclusion}\label{sec:Conclusion} We address real-time
freight bundling on OFEX platforms by formulating the multi-commodity one-to-one
pickup-and-delivery selective traveling salesperson problem (\emph{m1-PDSTSP}), a selective problem that couples combinatorial bundle selection with pickup-and-delivery routing under tight latency constraints.
Our algorithmic contribution is the learning--accelerated hybrid search pipeline, supported by attraction basin analysis and empirical evidence, that clarifies how neural construction, neighborhood-based improvement, and multi-start diversification can be effectively integrated for selective routing problems. We show that Transformer-based policy-gradient constructors can consistently generate solutions that lie within high-quality attraction basins, providing effective initializations for downstream neighborhood search. Building on this insight, we propose MSLNS, a multi-start metaheuristic with softmax-biased removal and adaptive destroy sizes, which systematically exploits basin structure to balance diversification and intensification.
Extensive experiments demonstrate that this combination advances the quality–-time frontier of selective PDP algorithms: compared to classical metaheuristics and standalone neural methods, the proposed hybrid pipeline closes optimality gaps more reliably under strict time budgets and approaches exact-solver performance on small instances while scaling effectively to larger ones. Beyond the specific \textit{m1-PDSTSP}, the evidence supports a broader algorithmic principle: high-quality learned initializations, when combined with basin-aware neighborhood exploration, can substantially enhance the effectiveness of improvement heuristics for selective TSP and PDP variants.

A key limitation is our reliance on rolling-horizon snapshots with fixed
per-snapshot budgets: under bursty arrivals or very large instances, latency
escalates and the method's advantage over simpler heuristics diminishes, thereby
limiting deployability at scale. Future work includes scaling our hybrid
pipeline via hierarchical (e.g., region) decomposition and distributed LNS, enabling parallel search and preserving sub-second response time even on very large snapshots. 
We also plan to validate the algorithm on real-world OFEX instances with additional operational constraints (e.g., time windows), quantifying latency, stability, and improvements in utilization and revenue.




\bibliographystyle{informs2014} 
\bibliography{sample} 

{This publication is part of the project FMaaS - it's a cargo match (with project number NWA.1518.22.023) of the research programme Dutch Research Agenda - research along NWA routes by consortia which is (partly) financed by the Dutch Research Council (NWO).}
%
\newpage
\section*{Appendices}

\section{Additional Methodology Details} \label{appendix:attention_model}
\subsection{Mixed Integer Linear Programming} \label{appendix:MILP}
We formulate a MILP that maximizes total collected revenue. The model explicitly
encodes precedence, capacity, and route-length constraints and supports exact
solution and bounding methods. Solving the MILP provides a ground-truth reference for
evaluating heuristics on small instances.

\paragraph{Decision variables:}
For every two distinct vertices $i, j \in\mathcal{V}$, let $X_{ij}\in\{0,1\}$ indicate
whether edge $(i,j)$ is visited, let $i_h$ denote the pickup vertex of request
$h\in \mathcal{H}$, let $T_i\ge 0$ denote the cumulative length from vehicle's start vertex $0$ to vertex $i$ along
the selected route, and $L_i\ge 0$ the load on departure from vertex $i$.

\resizebox{0.9\linewidth}{!}{%
  \begin{minipage}{\linewidth}
\begin{align} 
    \small
    \min \quad &
    - \sum_{h\in \mathcal{H}} r_h \Big(\sum_{j\in V} X_{i_h j}\Big)
    \label{eq:obj2}
    \\[4pt]
    \text{s.t.}\quad
    & X_{ii} = 0, & \forall i\in V,
    \label{eq:noself2}\\
    & \sum_{j\in \mathcal{P}} X_{0j} = 1,
    \quad
    \sum_{i\in \mathcal{D}} X_{i,\,2n+1} = 1,
    \label{eq:startend2}\\
    & \sum_{i\in V} X_{i0} = 0,
    \quad
    \sum_{j\in V} X_{2n+1,\,j} = 0,
    \label{eq:noinoutSE}\\
    & \sum_{j\in V} X_{ij} = \sum_{j\in V} X_{ji}, & \forall i \in \mathcal{P}\cup D,
    \label{eq:flowcons2}\\
    & \sum_{j\in V} X_{ij} \le 1, & \forall i \in V\setminus\{2n+1\},
    \label{eq:outdeg2}\\
    & \sum_{i\in V} X_{ij} \le 1, & \forall j \in V\setminus\{0\},
    \label{eq:indeg2}\\
    & \sum_{j\in V} X_{ij} \;=\; \sum_{j\in V} X_{j,\,i+n}, & \forall i \in \mathcal{P},
    \label{eq:pairvisit2}\\
    & T_0 = 0,\qquad T_{2n+1} \le T,
    \label{eq:routeLen2}\\
    & X_{ij}=1 \;\Rightarrow\; T_j \;\ge\; T_i + c_{ij}, & \forall i,j \in V,\; i\ne j,
    \label{eq:compatDist}\\
    & \sum_{j\in V} X_{ij}=1 \;\Rightarrow\; T_{i+n} \;\ge\; T_i, & \forall i\in \mathcal{P},
    \label{eq:precedence2}\\
    & L_0 = 0,\quad 0 \le L_i \le Q, & \forall i\in V,
    \label{eq:capbounds2}\\
    & X_{ij}=1 \;\Rightarrow\; L_j = L_i + q_j, & \forall i\in V,\; j\in \mathcal{P},
    \label{eq:loadPick}\\
    & X_{ij}=1 \;\Rightarrow\; L_j = L_i - q_{j-n}, & \forall i\in V,\; j\in \mathcal{D},
    \label{eq:loadDel}\\
    & L_{2n+1} = 0,
    \label{eq:endzero}\\
    & X_{ij}\in\{0,1\}, & \forall i,j\in V,\; i\ne j,
    \label{eq:binary2}\\
    & T_i \ge 0,\; L_i \ge 0, & \forall i\in V.
    \label{eq:contvars2}
\end{align}
\end{minipage}%
}
\medskip

\paragraph{Constraint explanations:}
Objective \eqref{eq:obj2} minimizes the negative collected revenue. Start/end
anchoring \eqref{eq:startend2}--\eqref{eq:noinoutSE} and flow/selectivity
\eqref{eq:flowcons2}--\eqref{eq:indeg2} enforce a single path that visits each
vertex at most once. Pairing \eqref{eq:pairvisit2} and precedence
\eqref{eq:precedence2} ensure a request is served if and only if both pickup and delivery
are included and the pickup precedes the delivery (in cumulative-length
order). Length-compatibility \eqref{eq:compatDist} ties visited arcs to
cumulative length and, with positive arc lengths, eliminates subtours;
\eqref{eq:routeLen2} caps total route length. Capacity bounds
\eqref{eq:capbounds2} with load propagation
\eqref{eq:loadPick}--\eqref{eq:loadDel} maintain feasible loads;
\eqref{eq:endzero}--\eqref{eq:contvars2} set terminal load
and variable domains.

\subsection{Attention Mechanism} 
In the encoder, each node's hidden state $\mathbf{h}_i$ is first projected into three distinct representations: queries $\boldsymbol{q}_i$, keys $\boldsymbol{k}_i$ and values 
$\boldsymbol{v}_i$ by multiplying with learnable weight matrices $W^Q$, $W^K$ and $W^V$. The query and key projections both map the original embedding (dimension $d_{\mathbf{h}}$) into a $d_k$-dimensional space, while the value projection maps into a $d_v$-dimensional space. Intuitively, 
$\boldsymbol{q}_i$ represents what node $i$ “asks” of its neighbors, $\boldsymbol{k}_j$ represents how node $j$ “presents” itself to be attended, and $\boldsymbol{v}_j$ is the actual information node $j$ contributes once its key matches a query.
We define dimensions $d_k$ and $d_v$ and compute the query $\boldsymbol{q}_i \in \mathbb{R}^{d_k}$, key $\boldsymbol{k}_i \in \mathbb{R}^{d_k}$, value $\boldsymbol{v}_i \in \mathbb{R}^{d_k}$ for each node by projecting the embedding $\mathbf{h}_i^{(\ell)}$ for all $\ell \in \{1, \dots, L\}$ as 
\(
    \boldsymbol{q}_{(c)}^{(\ell)} = W^Q \mathbf{h}_{(c)}^{(\ell)}, \, \boldsymbol{k}_i^{(\ell)} = W^K \mathbf{h}_i^{(\ell)},  \, \boldsymbol{v}_i^{(\ell)} = W^V \mathbf{h}_i^{(\ell)} .
\)
Here parameters $W^Q$ and $W^K$ are $d_k \times d_{\mathbf{h}}$ matrices and $W^V$ is a $d_v \times d_{\mathbf{h}}$ matrix. 

When computing the updated representation for node $i$, we form the
compatibility between $\boldsymbol{q}_i$ and each neighbor's key $
\boldsymbol{k}_j$ via their dot product, scaled by $1/\sqrt{d_k}$ to prevent the
magnitudes from growing too large as dimensionality increases. These scores are
optionally masked and then passed through a softmax function across all nodes $j
\in V$, yielding a distribution of attention weights.
\begin{equation}
    \mathbf{h}_{i}' = \sum_{j}\text{softmax}\bigg(\frac{{\boldsymbol{q}_i^{(\ell)}}^\top \boldsymbol{k}_j^{(\ell)}}{\sqrt{d_k}} \odot \mathcal{M}_t\bigg) \, \boldsymbol{v}_j^{(\ell)},
\label{eq:compatibility}
\end{equation}
where $d_k = \frac{d_{\mathbf{h}}}{M}$ is the query and key dimensionality,
$\odot$ denotes the Hadamard (element-wise) product and $\mathcal{M}_t$ is the
mask matrix. The new vector $\mathbf{h}_i'$ is simply the weighted sum of the
neighbors' value vectors, so that nodes with keys most aligned to the query of
node $i$ contribute more heavily to its updated embedding.

\paragraph{Multi-Head Attention:} 
To capture multiple types of interactions simultaneously, the model splits the
projections into $M$ separate heads. In each head $m$, its own triplet of
weight matrices $W_{m}^{Q}$, $W_{m}^{K}$ and $W_{m}^{V}$ computes a
head-specific output $\mathbf{h}_{im}'$ with $d_k = d_v =
\frac{d_{\mathbf{h}}}{M} = 16$. These $M=8$ head outputs are then linearly
projected back to the original hidden size $d_{\mathbf{h}}$ via head-specific
matrices $W_m^O$, and summed. This multi-head architecture allows different
subspaces to specialize, for instance one head focusing on content information
while another captures positional relationships.
\begin{equation}
    \textbf{MHA}_i (\mathbf{h}_1, \dots, \mathbf{h}_n) = \sum_{m=1}^{M} W_{m}^{O} \mathbf{h}_{im}' .
\end{equation}

\paragraph{Feed-Forward Sublayer:}
Following the attention mechanism, each node's representation passes through a
two-layer feed-forward network applied independently to each position. The first
linear layer expands the dimension from $d_{\mathbf{h}}$ up to a larger hidden
size $d_{\text{ff}}=512$, applies a ReLU activation function, and consequently projects back
down to $d_{\mathbf{h}}$. This position-wise MLP enriches each node's features
with non‑linear combinations of the attention output, enabling the network to
model complex interactions that attention alone cannot capture.
\begin{equation}
    \textbf{FF} (\hat{\mathbf{h}}_i) = W^{\text{ff},1} \cdot \text{ReLu} (W^{\text{ff},0} \hat{\mathbf{h}}_i + \mathbf{b}^{\text{ff},0}) = \mathbf{b}^{\text{ff},1} . 
\end{equation}

\paragraph{Batch Normalization}
Batch normalization is applied to stabilize training. For each feature dimension, the activations across the batch are normalized to zero mean and unit variance, then rescaled and shifted by learned parameters $w^{\text{bn}}$ and $b^{\text{bn}}$. This normalization occurs after the feed‑forward sublayer, smoothing out differences in scale across training examples and allowing for higher learning rates.
\begin{equation}
    \textbf{BN}(\mathbf{h}_i) = w^{\text{bn}} \odot \overline{\textbf{BN}} (\mathbf{h}_i) + \mathbf{b}^{\text{bn}},
\end{equation}
where $\overline{\textbf{BN}}$ refers to batch normalization without affine transformation.

\subsection{Traing Algorithm} \label{appendix:training_algorithm}
Given an instance $s = (f,\mathcal{F}) \in \mathcal{S}_\mathcal{X}$, the
objective is to find the optimal solution for $\pi^* = \arg
\min_{\pi \in \mathcal{F}} f(x)$. 
Probabilistic COP solvers tackle instance problems by parameterizing the solution
space with a continuous and differentiable vector $\boldsymbol{\theta}$ and
defining $p_{\boldsymbol{\theta}}(\pi|s)$  a parameterized conditional
distribution over discrete feasible complete solutions $\pi \in \mathcal{F}$. The
goal is to
minimize the expected negative collected revenues under $p_{\boldsymbol{\theta}}$ given by 
\begin{equation}
    \mathcal{L}(\boldsymbol{\theta}|s) = \mathbb{E}_{\pi \sim p_{\boldsymbol{\theta}} (\cdot|s)} [f(\pi)] = \sum_{\pi \in \mathcal{F}} p_{\boldsymbol{\theta}}(\pi|s)f(\pi).
    \label{eq:expected_cost}
\end{equation}

Reinforcement learning (RL) algorithms can be applied as a gradient estimator to
optimize such probabilistic generative models. The gradient of (\ref{eq:expected_cost}) can be formulated
using the well-known REINFORCE algorithm:
\begin{equation} 
    \scalebox{0.9}{$
    \nabla_{\boldsymbol{\theta}}\mathcal{L}(\boldsymbol{\theta}|s) = \mathbb{E}_{\pi \sim p_{\boldsymbol{\theta}} (\cdot|s)} [(f(\pi) - b(s))\nabla_{\boldsymbol{\theta}} \log p_{\boldsymbol{\theta}} (\pi|s)],
    \label{eq:reinforce_gradient}
    $}
\end{equation} 
where $b(s)$ denotes a baseline function that does not depend on $\pi$ and
estimates the expected negative collected revenues to reduce the variance of the
gradients. This estimator is unbiased and does not require
differentiable dynamics or rewards, making it suitable for discrete,
non-differentiable combinatorial decisions.

During training, the problem instances are independently and identically
distributed (i.i.d.) drawn from a distribution $\mathbb{S}$, so that the total
training objectives involves sampling from the distribution of instances, i.e.,
$\mathcal{L}(\boldsymbol{\theta}) = \mathbb{E}_{s \in
\mathbb{S}}\mathcal{L}(\boldsymbol{\theta}|s)$. By drawing $B$ i.i.d. sample
instances $s_1, s_2, \dots, s_B \sim \mathbb{S}$ and sampling a single feasible
complete solution per instance, i.e., $\pi_i \sim p_{\boldsymbol{\theta}}
(\cdot|s_i)$, the gradient in (\ref{eq:reinforce_gradient}) can be approximated
via Monte Carlo sampling (average the score-function terms):
\begin{equation}
    \scalebox{0.9}{$
    \nabla_{\boldsymbol{\theta}}\mathcal{L}(\boldsymbol{\theta}|s) \approx \frac{1}{B}\sum_{i=1}^{B} \bigg(f(\pi_i) - b(s_i) \bigg)\nabla_{\boldsymbol{\theta}} \log p_{\boldsymbol{\theta}} (\pi_i|s_i),
    \label{eq:monte-carlo sampling reinforce}
    $}
\end{equation}
The whole training procedure of AM \citep{kool2018attention} is summarized in Alg.~\ref{alg:reinforce}.
\vspace{-0.3cm}
\begin{algorithm}
\caption{REINFORCE with Rollout Baseline}
\footnotesize
\begin{algorithmic}[1]
\State \textbf{Input}: number of epochs $E$, steps per epoch $T$, batch size $B$, significance $\alpha$
\State Initialize $\boldsymbol{\theta}$, $\boldsymbol{\theta}^{\text{BL}} \leftarrow \boldsymbol{\theta}$ 
\For{epoch $= 1, \dots, E$}
    \For{step $= 1, \dots, T$}
        \State $s_i \leftarrow \textsc{RandomInstance}() \quad \forall i \in \{1, \dots, B\}$
        \State $\pi_i \leftarrow \textsc{SampleRollout}(s_i, p_{\boldsymbol{\theta}}) \quad \forall i \in \{1, \dots, B\}$  
        \State $\pi_i^{\text{BL}} \leftarrow \textsc{GreedyRollout}(s_i, p_{\boldsymbol{\theta}}^{\text{BL}}) \quad \forall i \in \{1, \dots, B\}$  
        \State $\nabla \mathcal{L} \leftarrow \frac{1}{B}\sum_{i=1}^{B} \Big(f({\pi}_i) - f(\pi_i^{\text{BL}})\Big)\nabla_{\boldsymbol{\theta}} \log p_{\boldsymbol{\theta}} (\pi_i)$
        \State $\boldsymbol{\theta} \leftarrow \textsc{Adam} (\boldsymbol{\theta}, \nabla \mathcal{L})$ 
    \EndFor
    \If{$\textsc{OneSidedPairedTTest}(p_{\boldsymbol{\theta}}, p_{\boldsymbol{\theta}}^{\text{BL}})<\alpha$ } $\boldsymbol{\theta}^{\text{BL}} \leftarrow \boldsymbol{\theta}$ 
    \EndIf
\EndFor
\end{algorithmic}
\label{alg:reinforce}
\end{algorithm}

\subsection{Definition of Solution Space} \label{appendix:solution_space}
Let $\mathcal{X}$ denote the set of all possible and feasible \textit{partial
solutions} for a given COP, $\mathcal{F} \in \mathcal{X}$. Let $\mathcal{Z}$ be
the set of \emph{elementary construction steps} (adding one pickup vertex, one
delivery vertex or the end vertex $z^e$). A solution is a finite sequence of
steps. We assume that $\mathcal{X}$ is equipped with a concatenation operator $
\circ$ (for a sequence $x = z_1 \circ \dots \circ z_k$ and a step $z\in Z$, $x \circ z := z_1 \circ \dots \circ z_k \circ z$) having a neutral element $\epsilon$ (empty sequence) such that $\epsilon
\circ x = x$. For a sequence $\boldsymbol{z}_{1:k}= z_1 \circ \cdots \circ z_k$
with $k\geq 1$ and \(\boldsymbol{z}_{1:k}\in\mathcal{Z}^k\), the set of partial
solutions is \(\mathcal{X} = \{\boldsymbol{z}_{1:k} :
C(\boldsymbol{z}_{1:\ell})=\mathrm{T}\;\;\forall \ell\le k \}. \) For any
$x\in\mathcal{X}$, we define its \emph{admissible step set} as
$$
A(x) := \{ z \in \mathcal{Z} : C(x\circ z)=\mathrm{T} \}.
$$
A partial solution $x\in\mathcal{X}$ is \emph{complete} if and only if no admissible step remains,
\[
x \text{ is complete } \iff A(x)=\emptyset.
\]
We define \((\mathcal{X}, \circ, \epsilon, \mathcal{Z})\) as the solution space.

For the \textit{m1-PDSTSP}, operator $\circ$ is the sequence concatenation,
$\epsilon$ is the empty sequence including trivial route at the vehicle's start
vertex $0$, and the step is a sequence of length $1$. The predicate
$C(\boldsymbol{z}_{1:\ell})=\mathrm{T}$, where $\boldsymbol{z}_{1:\ell} \in \mathcal{Z}^\ell$, is interpreted as a \emph{closure guarantee}: there
exists at least one finite extension that serves all pending deliveries and
terminates at the end vertex $z^{e}=2n{+}1$. Formally,
\[
    C(\boldsymbol{z}_{1:\ell})=\mathrm{T} \;\Longrightarrow\; \exists\, y\in \mathcal{Z}^\ast\ \text{s. t.}\ \boldsymbol{z}_{1:\ell} \circ y \circ z^e \in \mathcal{F},
\]    
where $y$ is a concatenation of the outstanding delivery steps (in some order)
and $\mathcal{Z}^\ast$ denotes the set of all finite sequences of all
outstanding delivery steps. A partial solution $x$ is feasible and
complete if and only if it reaches vertex $2n{+}1$ while satisfying all
operational constraints (capacity, route-length, and pickup-delivery
precedence) at every prefix. Consequently, any feasible, complete solution is a
composition of admissible steps starting from $\epsilon$ (vertex $0$) and ending at vertex
$2n{+}1$. The problem thus reduces to selecting a sequence $(z_1,\dots,z_\tau=z^e)$
with $z_\ell \in A(\boldsymbol{z}_{1:\ell-1})$ for all $\ell \leq \tau$, such that
$\boldsymbol{z}_{1:\tau} \in \arg\min_{x\in \mathcal{F}} f(x)$, a task which can
naturally be modeled as an Markov decision process (MDP).

\subsection{Markov Decision Process for Improvers} \label{appendix:mdp_improvers}
The MDP for improvers has the same solution space $\mathcal{F}$ but different
state space $\bar{\mathcal{X}} := x \in \mathcal{F}$, action space $\mathcal{A}
:= \{a \in A(x)\} \cup \{\epsilon\}$, where $A(x)$ is the set of admissible step
such that $x \xrightarrow{z} x'$ for all $z \in A(x)$ where $x' \in \mathcal{F}$
and $|x|<|x'|$. The labelled transition system is $x \xrightarrow[z]{f(x)-f(x')}
x'$ and the reward function is $r(x,z)=f(x)-f(x')$.

\section{Details of Baselines} \label{appendix:baselines}
We compare our MSLNS algorithm with several baselines. Tab.~\ref{tab:inference_baselines}
introduces the baseline decoders first, describing their main strategies and
hyperparameters, and lists the combinatorial
optimization (CO) post-improvement operators that can be applied on top of a
decoder, together with their configurations. We also provide the computational complexity of all baselines and improvement operators in Tab.~\ref{tab:inference_baselines}.

Since the effectiveness of different decoders (e.g., BS, MS,
sampling) is largely determined by the number of candidate trajectories
explored, we equalize their hyperparameters (e.g., setting $\beta = M = n/2$) to
align computational budgets and prevent artificial advantages. Similarly, when
comparing improvers (BI-LNS, LNS, MSLNS, SA), which may not converge within a fixed time
budget, we equalize their time budgets (e.g., $t_{\max}=1$s) to ensure a fair
comparison. 

\begin{table}[h!]
    \centering
\caption{Baselines descriptions}
\label{tab:inference_baselines}
\setlength{\tabcolsep}{2.5pt}      
\renewcommand{\arraystretch}{1.05} 
\resizebox{0.95\columnwidth}{!}{
\begin{tabular}{p{3cm}||p{9cm}|p{7cm}|p{5cm}}
    \hline
    Algorithm & Description & Hyperparameters & Complexity \\
    \hline
    \multicolumn{4}{c}{Baselines} \\
    \hline
    AM   & Greedy decoding strategy of the AM & - & $\mathcal{O}(n^2 L)$ \\
    AM-sample & Sampling decoding strategy of the AM & $\eta\in\{n/2,100\}$ samples & $\mathcal{O}(\eta \,n^2 L)$  \\
    AM-BS & Beam Search decoding strategy of the AM & beam width $\beta \in \{n/2, 100\}$ & $\mathcal{O}(\beta \,n^2 L)$ \\
    AM-MS & Multi-Start greedy trajectories of the AM & $M=n/2$ & $\mathcal{O}(M \,n^2 L)$\\
    POMO-single & Single greedy trajectory of POMO & - & $\mathcal{O}(n^2 L)$ \\
    POMO      & Multi greedy trajectories of POMO & $M=n/2$ & $\mathcal{O}(M \,n^2 L)$\\
    POMO-SGBS & SGBS \citep{choo2022simulation} combined with POMO & $\beta=\gamma=5$ & $\mathcal{O}(\beta \, \gamma\, n^2 L)$ \\
    Gurobi & Gurobi solver v12.0.3 & - & - \\
    GS & Greedy Search heuristic & - & $\mathcal{O}(n)$\\
    MSG & Multi-Start Greedy Search heuristic & $M=n/2$ & $\mathcal{O}(M\,n)$\\
    LNS & Large Neighborhood Search with random destroy & time out $t_{max}\in\{1s,4s,10s\}$ or iteration $I$ & $\mathcal{O}(I\,n^3)$ \\
    SA & Simulated Annealing & time out $t_{max}\in\{1s,4s,10s\}$ & -\\
    \hline
    \hline
    \multicolumn{4}{c}{Post-improvement CO operators} \\
    \hline
    2\text{-}Opt    & Local 2\text{-}Opt refinement operator & until no gain & $\mathcal{O}(n^2)$\\
    HC              & Hill Climbing on neighborhood & until no gain & $\mathcal{O}(n^3)$ \\
    BI-LNS          & 1-destroy Best-improvment LNS & time out $t_{\max}\!\in\!\{1\text{s},4\text{s},10\text{s}\}$ & $\mathcal{O}(n^4)$\\
    MSLNS         & Multi-Start Large Neighborhood Search & $M \in [5,10]$, $\beta\in\{3,5\}$, $t_{\max}\!\in\!\{1\text{s},4\text{s},10\text{s}\}$ & $\mathcal{O}(\lfloor{n}/{k_{max}}\rfloor\,M\,n^3)$  \\
    \hline
    \end{tabular}}
    
    \smallskip
\footnotesize{$L$ is the layer size of Transformer decoder. For baselines, we do not include masking costs in the complexity analysis.}
\end{table}

\paragraph{Simulation-guided Beam Search (SGBS)} is a beam search-based constructing
 algorithm proposed by \cite{choo2022simulation} that examines candidate
 solutions within a fixed-width tree search that both a DNN-learned policy and a
 simulation (rollout) identify as promising. Given that SGBS is designed for
 unselective routing problems, we customize it for the \textit{m1-PDSTSP}.
 Similar to MSLNS, after each step, before selecting the top $\beta$ candidate
 routes, we remove the duplicated routes (routes with same selected requests
 set) to ensure diversity. During the beam search, the rollout simulation is
 performed using the greedy decoding strategy of the POMO model as in
 Alg.~\ref{alg:pomo-inference}.

 \begin{algorithm}
\caption{Policy Optimization with Multiple Optima Inference}
\footnotesize
\begin{algorithmic}[1]
\State \textbf{Input}: instance $s$, policy $p_{{\theta}}$, number of start pickup nodes per sample $M$
    \State $\{\pi^1_{1}, \pi^2_{1}, \dots, \pi^M_{1} \} \leftarrow \text{SelectStartPickupNodes}(s)$
    \State $\pi^j \leftarrow \text{GreedyRollout}(\pi_{1}^j,s, p_{{\theta}}) \quad \forall j \in \{1,\dots,M\}$  
    \State $j_{min} \leftarrow \arg\min_{j \in \{1,\dots,M\}} f(\pi^j)$
    \State \textbf{return} $\pi^{j_{min}}$
\end{algorithmic}
\label{alg:pomo-inference}
\end{algorithm}

\paragraph{Greedy Search} is a constructor that iteratively selects the option
that offers the most immediate benefit at each step, without regard for future
consequences. We design the GS algorithm based on the problem structure. Each
new node is chosen based on fitness, which is defined as follow: 
\begin{equation}
    \text{fitness}_{j} = r_j - c_{i,j}, \quad \forall j \in \mathcal{V}.
    \label{eq:greedy_fitness}
\end{equation}

\paragraph{Multi-Start Greedy Search} is an extension of GS that
generates multiple initial solutions by varying the $M$ starting pickup nodes based
on (\ref{eq:greedy_fitness}), and selects the best among them.

\paragraph{Large Neighborhood Search} is an improver that iteratively destroys a
portion of the current solution and repairs it to explore a larger solution
neighborhood. To implement LNS for \textit{m1-PDSTSP}, we employ the random destroy
($k\sim\text{Unif}\{1,\dots,k_{max}\}$ requests drawn uniformly at random) and
greedy insertion repair strategies (Alg.~\ref{alg:repair}). The complexity of
\textsc{REPAIR} is $\mathcal{O}(n^3)$. Through our experiments, we find that
$k_{max}=3$ generates the best performance. 

\begin{algorithm}[t]
\caption{\textsc{Repair}/Greedy Insertion}
\footnotesize
\begin{algorithmic}[1]
\Require route $\pi^*$, instance size $n$, load list $q$, capacity limit $Q$, maximum length limit $T$, reward function $f$
\While{$\pi^*=\pi$}
\State $\pi\!\leftarrow\!\pi^*$
\State Compute remaining capacity $\textit{cap\_left}$ along route $\pi$
\State Initialize $\textit{ValidRoutes}\!\leftarrow\!\{\pi\}$, $\textit{Scores}\!\leftarrow\!\{R(\pi)\}$
\For{request $r=0$ to $n-1$ where $r\notin\pi$}
  \State $\textit{Intervals}\!\leftarrow\!\textsc{FindCapacityGap}(Q-\textit{cap\_left}-q[r])$
  \For{each interval $[s,e]\in\textit{Intervals}$}
    \State $\textit{candidates}\!\leftarrow\!\textsc{Insert}([s,e], r, r+n)$ \Comment{There can be multiple candidates}
    \For{each $\textit{cand}\in\textit{candidates}$}
      \If{Route length $c(\textit{cand})\!\le\!T$}
        \State Append $\textit{cand}$ to $\textit{ValidRoutes}$ and $R(\textit{cand})$ to $\textit{Scores}$
        \EndIf
    \EndFor
  \EndFor
\EndFor
\State $\pi^* \!\leftarrow\! \textit{ValidRoutes}[\arg\max \textit{Scores}]$\
\EndWhile
\State \Return $\pi^*$ 
\end{algorithmic}
\label{alg:repair}
\end{algorithm}

\paragraph{Simulated Annealing} is an improver that occasionally accepts worse
moves with probability $e^{-\Delta/\text{temperature}}$ under a cooling schedule
to escape local optima. To implement SA for the \textit{m1-PDSTSP}, traditional
operators like \textsc{INVERSE}, \textsc{SWAP}, and \textsc{INSERT$\_$SUBTOUR}
cannot be directly applied due to stringent constraints. Particularly,
\textsc{INSERT$\_$SUBTOUR} almost always violates nearly all constraints when a
subsequence is randomly repositioned. Because the route visits only a subset of
nodes, we treat \textsc{INSERT} as adding a new request (pickup-delivery pair)
rather than deleting and reinserting existing nodes.
As for the \textsc{INVERSE}
operator, its application is generally restricted to changing the order among
groups of consecutive pickup or delivery nodes. Effectively, this becomes a \textsc{SWAP}
operation when only two consecutive nodes are reordered. \textsc{SWAP}
also allows for the repositioning of a pickup node from one request with a
delivery node from another. Therefore, we can only keep the \textsc{SWAP} operator and
abandon the \textsc{INVERSE}.

In addition to adapting the INSERT and \textsc{SWAP} operators, a new operator called
\textsc{REPLACE} can be introduced. This operator allows for the replacement of the
pickup--delivery pair nodes of a selected request with those of an unselected
request in the same positions in the route. Furthermore, we can introduce a
\textsc{DESTROY} operator, which randomly removes a selected request from the route.
We set the hyperparameters as follows: initial temperature is 10, cooling rate is 0.95, stopping temperature is 0.01 and the number of iterations for each temperature is 20.


\subsection{Details of Improvers}

\paragraph{2-Opt} is an improver that repeatedly selects two edges in a tour, reverses the intervening segment (swapping the edges), and keeps the change if the new route is feasible and shortens the route, until no further improvement is possible.

\paragraph{Hill Climbing} is an improver that starts with an initial solution
and iteratively makes small changes to improve it. The algorithm examines
neighboring solutions and moves to a neighbor that offers an improvement (i.e., Alg.~\ref{alg:repair}) over the current one. This
process repeats until no further improvements can be found.

\paragraph{Best-improvement Large Neighborhood Search} \label{appendix:bilns} is an improver that starts from an initial solution,
iteratively enumerates all neighbors and picks the best, repeatedly moving to an
improving neighbor. The process continues until no improving move is found or a predefined stopping criterion is met. In this paper, we
implement a 1-destroy BI-LNS that selects only one requests in the current solution
to destroy and reinsert the destroyed solution greedily (Alg.~\ref{alg:repair}).
This reinsertion process repeats until no single-request reinsertion yields
improvement.




\section{Additional Revenue Settings} \label{appendix:additional_experiments}

\subsection{Revenue Settings}
We have experimented with four different revenue settings to evaluate the
robustness of our method. The details of the \textit{Ton-Distance}, \textit{Constant} and \textit{Uniform} settings are described below.

\begin{enumerate}
    \item \textbf{Ton-Distance}: This setting rewards requests in proportion to the
product of demand and pickup-delivery separation. $r_h = q_h \cdot c_{h,h+n} * \mu $, where
$mu\sim\text{Unif}(0.5,1.5)$ is the discount to add randomness, $\hat{r}_h
=\frac{r_h}{\max\{r_h\}}$. Ton-Distance is a standard
industry metric that reflects both the weight and distance of freight, aligning
with how carriers are often paid and how costs are incurred.
    \item \textbf{Constant}: Every node has the same revenue, therefore, the goal becomes to maximize
the number of served customers within the maximum length constraint. $r_h =
\hat{r}_h=1$. 

    \item \textbf{Uniform}: The revenue of each request is uniformly sampled from the set
$\{1,2,\dots,100\}$ .$r_h \sim \text{Uniform}(1,100)$, $\hat{r}_h =
\frac{r_h}{100}$. This setting introduces variability in revenues, reflecting
scenarios where different requests have different levels of importance or
urgency.
\end{enumerate}


\section{Additional Experimental Results}

\subsubsection{More details on Distance m1-PDSTSP} \label{appendix:more_details_distance}
\begin{figure}[h]
\includegraphics[width=\textwidth]{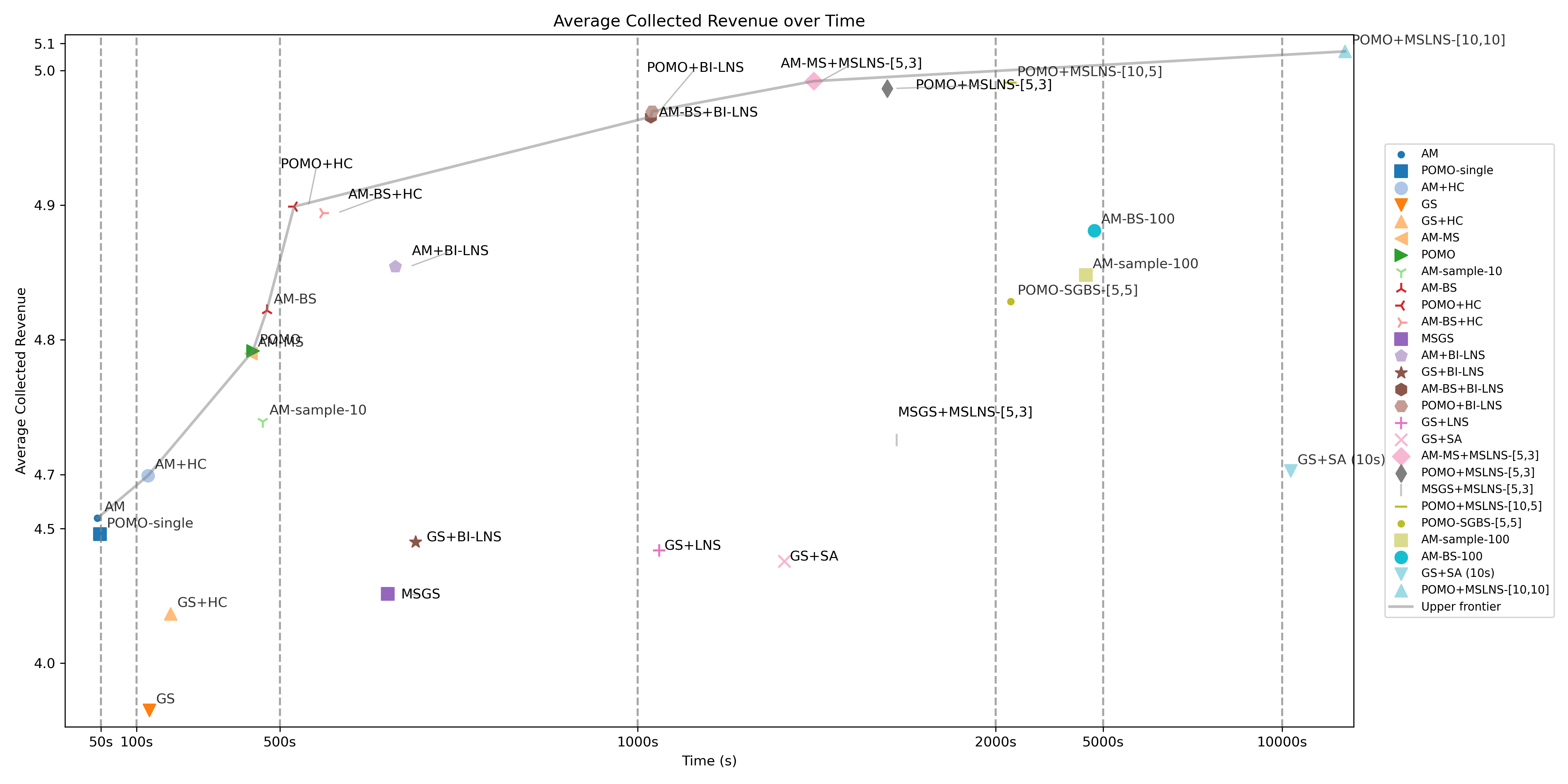}
\caption{Average collected revenue vs. wall-clock time for all contenders on
PDSTSP42. Each marker denotes a method; x-axis is runtime (vertical dashed lines
indicate time budgets) and y-axis is mean revenue. The Pareto front is traced by
AM/AM+HC (low), POMO+HC (middle), POMO/AM+MSLNS$_{[10,5]}^{t_{max}}$ and
POMO/AM-BS+BI-LNS$^{t_{max}}$ (high), and POMO+MSLNS$_{[10,10]}$ (higher).}
\label{fig:model_selection_42}
\end{figure}

Given AM, POMO and multiple baselines with numerous hyperparameters, we first
tune all methods on PDSTSP42 under the \emph{Distance} setting.
Hyperparameters are selected using a validation split, and the best methods
choice is made on the Pareto frontier of revenue (higher is better) versus
computational cost (lower is better) in Fig.~\ref{fig:model_selection_42}.


\begin{figure}
\centering
\begin{subfigure}{\textwidth}
    \includegraphics[width=\textwidth]{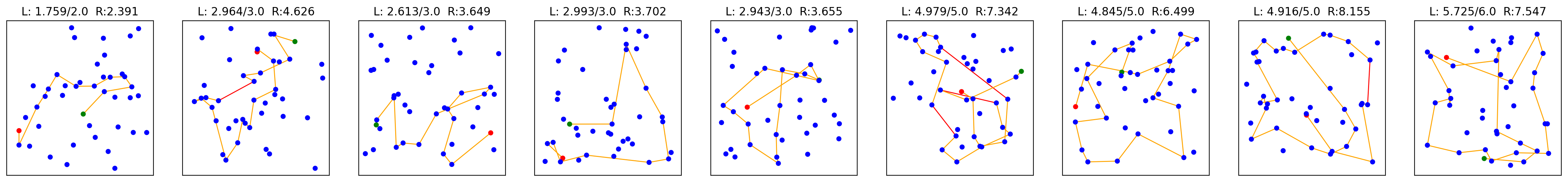}
\end{subfigure}
\begin{subfigure}{\textwidth}
    \includegraphics[width=\textwidth]{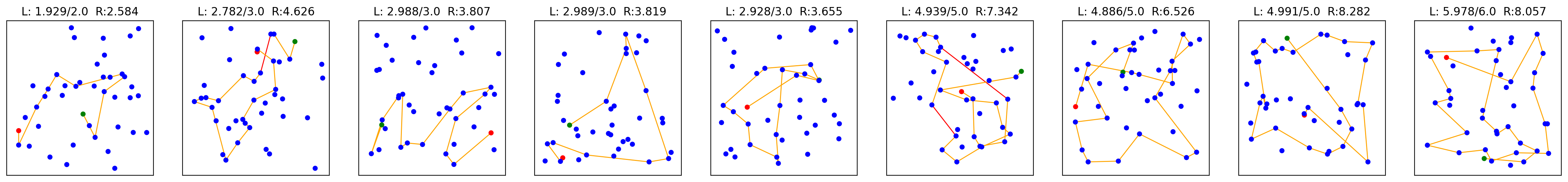}
\end{subfigure}
\begin{subfigure}{\textwidth}
    \includegraphics[width=\textwidth]{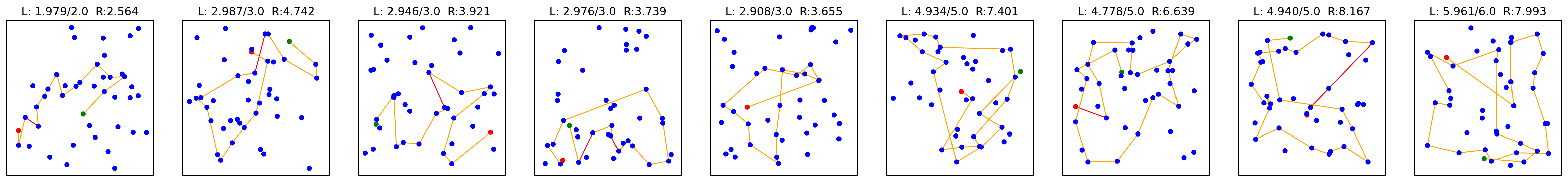}
\end{subfigure}
\begin{subfigure}{\textwidth}
    \includegraphics[width=\textwidth]{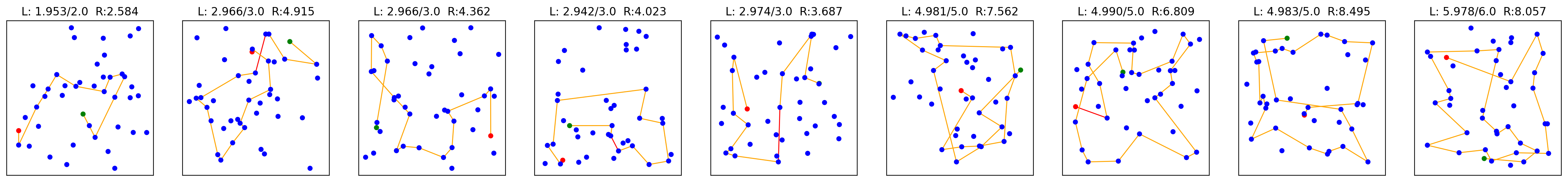}
\end{subfigure}
\caption{Example routes on \emph{Distance} PDSTSP42 with $Q$ and $T$ randomly
generated. Top to bottom: POMO, POMO+BI-LNS$^{t_{max}}$,
POMO+MSLNS$_{[10,5]}^{t_{max}}$, and POMO+MSLNS$_{[10,10]}$.}
\label{fig:example_routes}
\end{figure}

From Fig.~\ref{fig:model_selection_42}, a clear
Pareto front emerges. For small budgets ($\leq100s$), AM+HC dominates the
quick region ($\approx4.7$ avg revenue), outperforming GS, HC, AM, and
POMO-single. In the mid-budget range $500-1000s$, AM-BS improves a bit ($
\approx 4.8$) but is soon dominated by POMO+HC and AM-BS+HC. In the high-budget
range ($200s-5000s$), time-limited LNS-based and MSLNS-based hybrids give a
significant jump ($\approx 5.0$), beating LNS/SA and heavy sampling/beam-search/SGBS
variants. For higher budgets ($\geq 3000s$), POMO+MSLNS$_{[10,10]}$ is at the
top $(\approx5.1)$. Fig.~\ref{fig:example_routes} shows routes generated by
POMO, POMO+BI-LNS$^{t_{max}}$, MSLNS$_{[10,5]}^{t_{max}}$ and
POMO+MSLNS$_{[10,10]}$ for 9 random instances. 

\subsubsection{Different maximum route-lengths}\label{appendix:gap_size} We evaluate
the performance of different methods under various maximum length sizes $T$
for the \emph{Distance} setting to understand how route-length constraints
impact solution quality. We consider three sizes: tight ($T=2$), medium ($T=4$),
and large ($T=8$) with fixed capacity ($Q=10$), which correspond to different
levels of route-length tightness. For each size, we report the average revenue
achieved by each method on the test set of 1,000 instances. 

\newcommand{\vlabel}[1]{\rotatebox[origin=c]{90}{\scriptsize#1}}

\begin{table*}[t]
\centering
\caption{Performance matrix: PDSTSP42 vs.\ PDSTSP82 across gap classes (Tight/Middle/Wide) on the \emph{Distance} setting with $Q=10$. }
\label{tab:gap_results}
\small
\setlength{\tabcolsep}{3pt}
\renewcommand{\arraystretch}{0.95}
\resizebox{\textwidth}{!}{%
\begin{tabular*}{1.2\textwidth}{@{\extracolsep{\fill}}
    l
    l
    r r r r r
    r r r r r
@{}}
\toprule
 & & \multicolumn{5}{c}{\textbf{PDSTSP42}} & \multicolumn{5}{c}{\textbf{PDSTSP82}} \\
\cmidrule(lr){3-7}\cmidrule(lr){8-12}
\multicolumn{1}{c}{\textbf{}} & \multicolumn{1}{c}{\textbf{Algorithm}} &
\multicolumn{1}{c}{Time} & \multicolumn{1}{c}{Revenue} & \multicolumn{1}{c}{Length} & \multicolumn{1}{c}{Gap} & \multicolumn{1}{c}{WR} &
\multicolumn{1}{c}{Time} & \multicolumn{1}{c}{Revenue} & \multicolumn{1}{c}{Length} & \multicolumn{1}{c}{Gap} & \multicolumn{1}{c}{WR} \\
\midrule

\multirow{11}{*}{\vlabel{\textbf{Tight ($T=2$)}}}
& GS+BNS$^{t_{max}}$            & 2m  & 1.686 & 1.933 & 12.51\% & 32.1\% & 4m & 2.031 & 1.952 & 15.69\% & 18.4\% \\
& POMO-single                  & 9s  & 1.581 & 1.876 & 18.09\% & 14.2\% & 21s & 1.936 & 1.903 & 19.63\% & 5.5\%\\
& AM                           & 8s  & 1.668 & 1.877 & 13.45\% & 21.5\% & 27s & 2.039 & 1.902 & 15.36\% & 10.2\%\\
& AM-MS                        & 1m  & 1.815 & 1.911 & 5.80\%  & 39.7\% & 5m  & 2.265 & 1.932 & 5.98\% & 22.9\%\\
& POMO                         & 1m  & 1.820 & 1.896 & 5.49\%  & 40.8\% & 5m  & 2.263 & 1.935 & 6.06\% & 26.1\%\\
& AM-BS                        & 1m  & 1.847 & 1.917 & 4.03\%  & 53.2\% & 6m  & 2.315 & 1.942 & 3.90\% & 40.5\%\\
& POMO+BNS$^{t_{max}}$          & 3m  & 1.874 & 1.938 & 2.74\%  & 63.4\% & 10m & 2.335 & 1.944 & 3.07\% & 49.4\%\\
& AM-BS+BNS$^{t_{max}}$         & 3m  & 1.886 & 1.929 & 2.12\%  & 70.3\% & 11m & 2.356 & 1.948 & 2.20\% & 56.8\% \\
& POMO+MNS$_{[10,5]}^{t_{max}}$& 15m & \textbf{1.915} & 1.931 & \textbf{0.64\%}  & \textbf{87.3\%}  & 42m & \textbf{2.396} & 1.948 & \textbf{0.53\%} & \textbf{77.8\%} \\
& POMO+MNS$_{[10,10]}$         & 30m & 1.919 & 1.933 & 0.41\% & 91.8\% & 1.4h & 2.409 & 1.951 & 0.00\% & 89.6\% \\
& Gurobi                       & 20.0h & 1.927 & 1.956 & 0.00\% & 100.0\% & \multicolumn{5}{c}{--}\\       
\midrule
\multirow{10}{*}{\vlabel{\textbf{Middle ($T=4$)}}}
& GS+BNS$^{t_{max}}$            & 8m  & 4.285 & 3.948 & 11.46\%  & 8.3\%   & 22m & 5.058 & 3.965 & 15.29\% & 1.9\% \\
& POMO-single                  & 41s & 4.281 & 3.855 & 11.50\%  & 3.1\%   & 2m  & 5.178 & 3.906 & 13.28\% & 1.1\% \\
& AM                           & 37s & 4.329 & 3.857 & 10.53\%  & 4.4\%   & 2m  & 5.258 & 3.904 & 11.94\% & 1.8\% \\
& POMO                         & 8m  & 4.564 & 3.898 & 5.66\%   & 9.1\%   & 31m & 5.645 & 3.932 & 5.46\% & 5.6\% \\
& AM-MS                        & 6m  & 4.566 & 3.895 & 5.65\%   & 9.4\%   & 30m & 5.614 & 3.928 & 5.98\% & 6.3\% \\
& AM-BS                        & 6m  & 4.616 & 3.904 & 4.60\%   & 16.1\%  & 34m & 5.711 & 3.939 & 4.35\% & 13.9\% \\
& POMO+BNS$^{t_{max}}$          & 12m & 4.703 & 3.951 & 2.81\%   & 29.6\%  & 54m & 5.840 & 3.959 & 2.26\% & 29.6\% \\
& AM-BS+BNS$^{t_{max}}$         & 13m & 4.730 & 3.944 & 2.24\%   & 39.0\%  & 55m & 5.846 & 3.962 & 2.09\% & \textbf{37.6\%} \\
& POMO+MNS$_{[10,5]}^{t_{max}}$ & 31m & \textbf{4.766} & 3.930  & \textbf{1.52\%}  & \textbf{43.1\%}  & 1.5h  & \textbf{5.882} & 3.947 & \textbf{1.49\%} & 33.6\% \\
& POMO+MNS$_{[10,10]}$          & 1.3h  & 4.839 & 3.940 & 0.00\% & 75.0\%  & 3.7h & 5.971 & 3.953  & 0.00\% & 65.5\% \\
\midrule

\multirow{10}{*}{\vlabel{\textbf{Wide ($T=8$)}}}
& GS+BNS$^{t_{max}}$            & 18m & 8.062 & 7.954 & 11.19\% & 0.5\%  & 1.1h  & 9.850  & 7.963 & 17.04\%& 0.1\% \\
& POMO-single                  & 1m  & 8.389 & 7.818 & 7.60\%  & 1.1\%  & 4m  & 10.875 & 7.889 & 8.41\% & 0.1\% \\
& AM                           & 1m  & 8.416 & 7.811 & 7.30\%  & 0.7\%  & 4m  & 10.829 & 7.884 & 8.79\% & 0.3\% \\
& POMO                         & 10m & 8.747 & 7.860 & 3.69\%  & 2.9\%  & 1.2h  & 11.419 & 7.914 & 3.82\% & 0.8\% \\
& AM-MS                        & 12m & 8.759 & 7.845 & 3.50\%  & 6.0\%  & 1.2h  & 11.341 & 7.913 & 4.48\% & 2.8\% \\
& AM-BS                        & 13m & 8.766 & 7.866 & 3.43\%  & 4.6\%  & 1.2h  & 11.430 & 7.926 & 3.73\% & 2.1\% \\
& POMO+BNS$^{t_{max}}$          & 21m & 8.935 & 7.955 & 1.49\%  & 21.8\% & 2.2h  & \textbf{11.737} & 7.967 & \textbf{1.15\%} & 23.4\% \\
& AM-BS+BNS$^{t_{max}}$         & 25m & 8.979 & 7.947 & 1.09\%  & \textbf{35.3\%} & 2.2h  & 11.690 & 7.968 & 1.54\% & \textbf{28.4\%} \\
& POMO+MNS$_{[10,5]}^{t_{max}}$ & 32m & \textbf{8.989} & 7.920 & \textbf{0.98\%} & 24.7\% & 2.4h  & 11.719 & 7.951 & 1.30\% & 12.4\% \\
& POMO+MNS$_{[10,10]}$          & 3.0h  & 9.077 & 7.930 & 0.00\% & 53.3\% & 11.4h & 11.873 & 7.954 & 0.00\% & 50.5\% \\
\bottomrule
\end{tabular*}}

\smallskip
\emph{Notes:} WR: Winning Rate. MSN: MSLNS, BNS: BI-LNS. \\
\vspace{-0.2\baselineskip}
\end{table*}

Across both sizes (PDSTSP42 and PDSTSP82 in Tab.~\ref{tab:gap_results}) and all
gap classes (Tight/Middle/Wide), hybrid methods that start from NCO solutions
and apply search consistently dominate pure baselines. Single-shot baselines
(AM, POMO, and POMO-single) are clearly behind.
Adding search to
AM/POMO helps a lot: AM-BS, AM-MS, and POMO close much of the gap,
and adding improver give another solid boost. Except for the unlimited
POMO+MSLNS$_{[10,10]}$, its time-budgeted sibling
POMO+MSLNS$_{[10,5]}^{t_{\max}}$ is a practical substitute, reaching within
$\approx 0.5-1.5\%$ of the best with far less time. We see 
POMO+MSLNS$_{[10,5]}^{t_{\max}}$ is only surpassed by POMO+BI-LNS$^t_{max}$ on
wide PDSTSP82 as wide gaps on larger instances demand more destroy-repair time,
and the multi-start variant can hit the budget before fully exploring. Finally,
the WR columns on PDSTSP82 suggest AM-BS families often
win on Middle/Wide, indicating AM-BS and POMO explore complementary regions of the
solution space.

\subsubsection{Generalization experiments}
We assess out-of-distribution (OOD) generalization by training (and tuning
hyperparameters) on a \emph{source} distribution and evaluating \emph{without
further training} on a \emph{target} distribution that differs in problem
factors. Concretely, we vary instance scale $n$, models trained at PDSTSP42
are evaluated at PDSTSP82 and PDSTSP122, and models trained at PDSTSP82 are
evaluated at PDSTSP122. We report zero-shot performance, which demonstrates how well our
model performs on a new task or distribution without any additional training on that
target. This protocol quantifies robustness to scale/distribution shift, helps
detect overfitting, and guides design choices. In practice, problem scale and
mix evolve over time (new regions, more customers, seasonal peaks), and
retraining for every shift is rarely feasible; zero-shot evaluation thus
indicates how reliable and robust a method is.

\setlength{\arrayrulewidth}{0.6pt} 
\newcommand{\midsep}{0.6pt}        

\begin{table*}[t]
\centering
\caption{Performance matrix: PDSTSP82 vs.\ PDSTSP122 (Generalization from PDSTSP42) vs.\ PDSTSP122 (Generalization from PDSTSP82) on the \emph{Distance} setting with $Q$ and $T$ randomly generated.}
\label{tab:generalization_results}
\small
\setlength{\tabcolsep}{3pt}

\begin{tabular*}{\textwidth}{@{\extracolsep{\fill}}
  l
  r r r r               
  r r r r               
  !{\vrule width \arrayrulewidth} 
  r r r r               
@{}}

\Xhline{\midsep}

& \multicolumn{4}{c}{\textbf{PDSTSP82} (42)}
& \multicolumn{4}{c!{\vrule width \arrayrulewidth}}{\textbf{PDSTSP122} (42)}
& \multicolumn{4}{c}{\textbf{PDSTSP122} (82)} \\
\cline{2-5}\cline{6-9}\cline{10-13}

\multicolumn{1}{c}{\textbf{Algorithm}} &
\multicolumn{1}{c}{Time} & \multicolumn{1}{c}{Revenue} & \multicolumn{1}{c}{Profit} & \multicolumn{1}{c}{Gap} &
\multicolumn{1}{c}{Time} & \multicolumn{1}{c}{Revenue} & \multicolumn{1}{c}{Profit} & \multicolumn{1}{c!{\vrule width \arrayrulewidth}}{Gap } &
\multicolumn{1}{c}{Time} & \multicolumn{1}{c}{Revenue} & \multicolumn{1}{c}{Profit} & \multicolumn{1}{c}{Gap} \\
\Xhline{\midsep}

POMO-single          & 7m & 9.54 & 2.55 & 14.29\% & 18m & 13.92 & 3.23 & 22.80\% & 17m & 16.51 & 6.00 & 8.43\% \\
AM                   & 6m & 9.60 & 2.64 & 13.75\% & 18m & 14.86 & 4.18 & 17.48\% & 13m & 16.54 & 6.04 & 8.26\% \\
AM-MS                & 1.1h & 10.23 & 3.24 & 8.09\% & 2.9h & 15.32 & 4.65 & 15.03\% & 2.4h & 17.16 & 6.66 & 4.83\% \\
AM-BS                & 1.1h & 10.15 & 3.16 & 8.81\% & 2.9h & 15.48 & 4.80 & 14.14\% & 2.9h & 17.07 & 6.58 & 5.32\% \\
POMO                 & 1.1h & 10.20 & 3.21 & 8.36\% & 2.9h & 14.95 & 4.26 & 17.08\% & 2.0h & 17.25 & 6.75 & 4.33\% \\

\Xhline{\midsep}
\end{tabular*}
\end{table*}

Based on the results in Tab.~\ref{tab:generalization_results} and
Tab.~\ref{tab:main_results_distance}, we observe that all methods experience a
performance drop when generalizing to larger instances (except for AM trained on
PDSTSP82), which is expected due to the increased complexity and size of the
problem. Models trained on PDSTSP42 suffer a pronounced drop when evaluated on
PDSTSP82/122, whereas models trained on PDSTSP82 exhibit only minor degradation,
indicating that both AM and POMO learn more scale-robust policies once exposed
to mid-size instances. Note that this advantage is partly attributable to
greater compute: PDSTSP82 models were trained for roughly $2\times$ the total time
compared with PDSTSP42. Finally, POMO, AM-MS, and AM-BS effectively narrow the generalization gap by adapting
candidates at test time.

\begin{figure}
\centering
\begin{subfigure}{0.48\textwidth}
    \includegraphics[width=\textwidth]{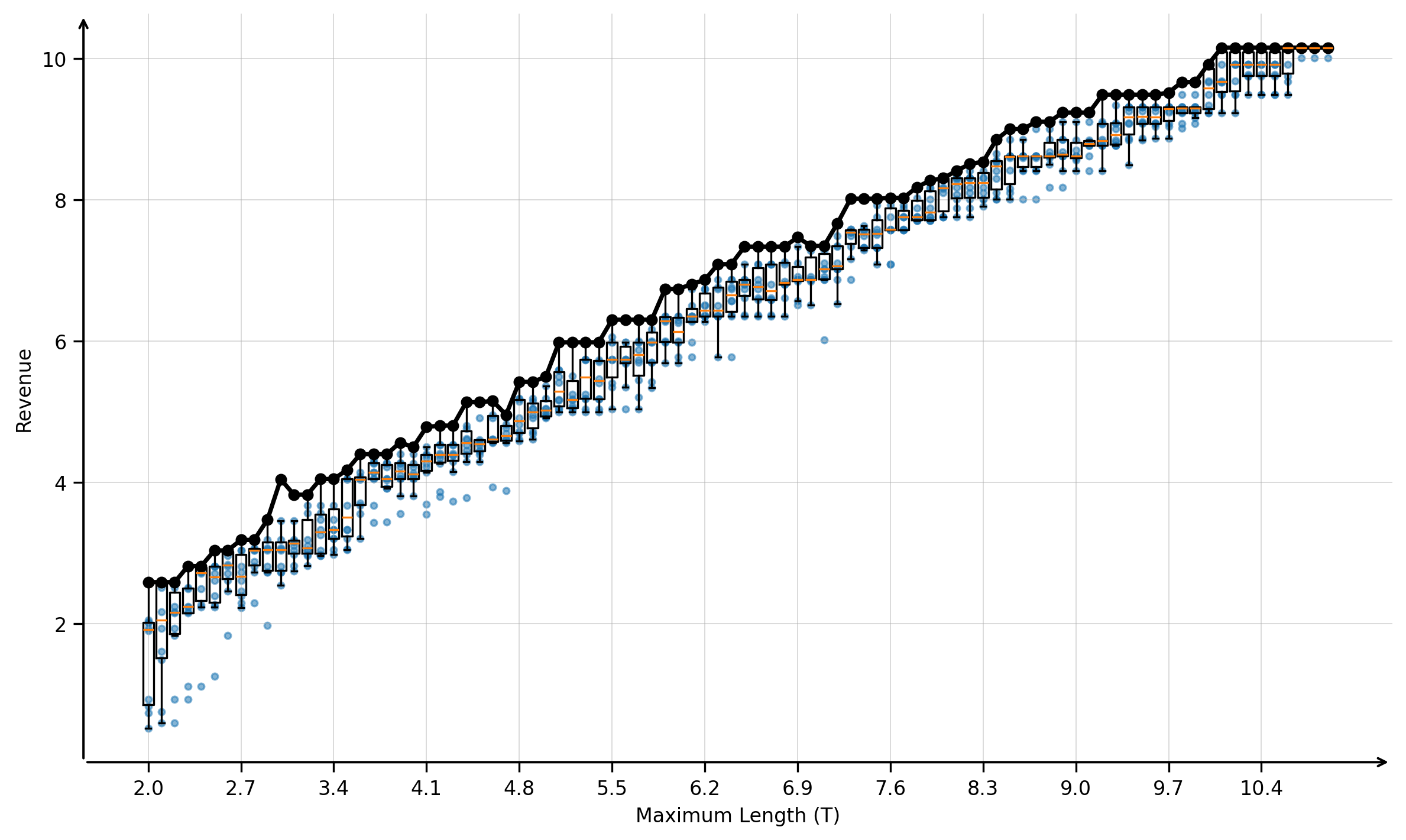}
    \caption{Zero-shot generalization across $T$ ($Q=10$)}
\end{subfigure}
\begin{subfigure}{0.48\textwidth}
    \includegraphics[width=\textwidth]{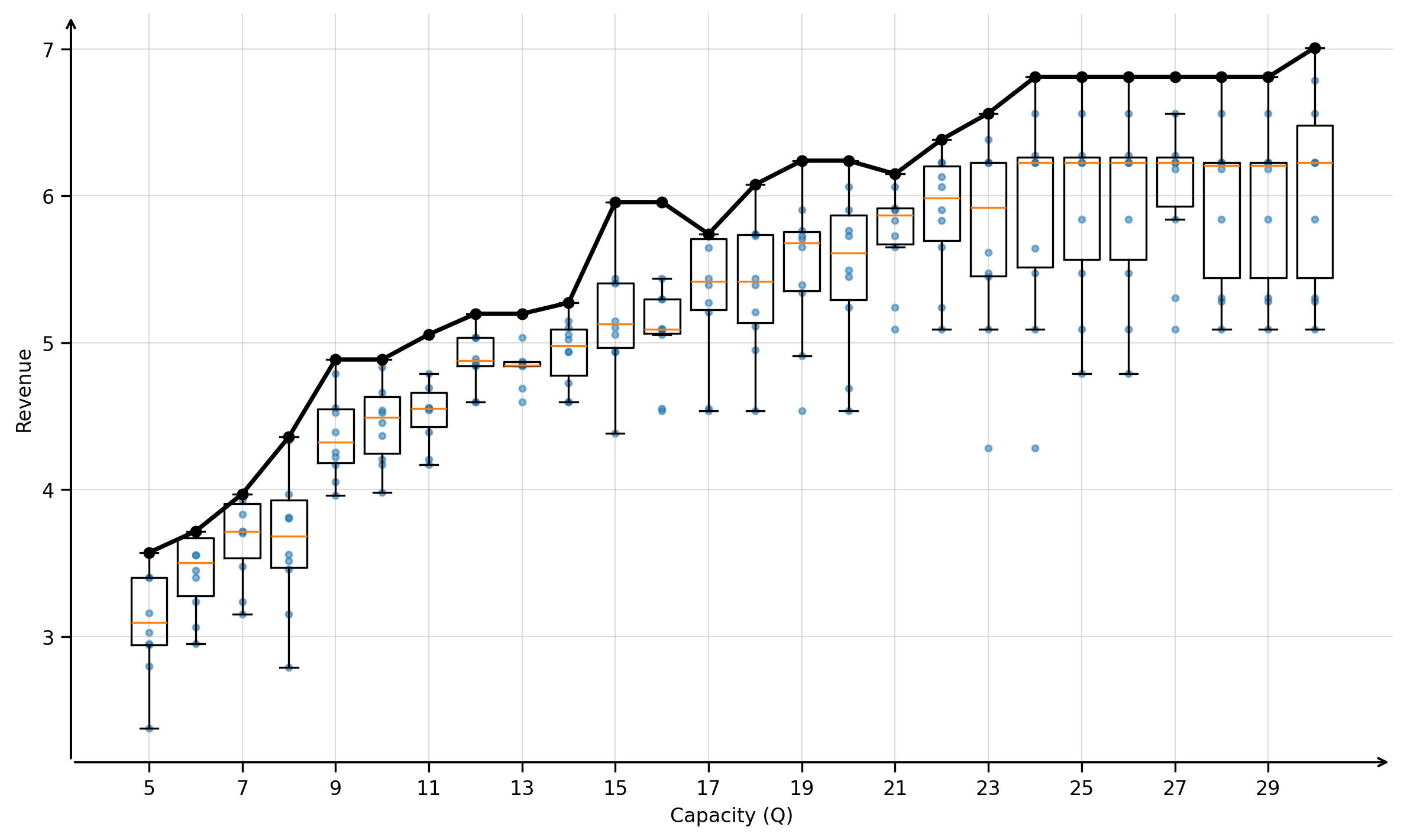}
    \caption{Zero-shot generalization across $Q$ ($T=4$)}
\end{subfigure}
\caption{Out-of-distribution robustness to constraint tightness (the route-length budget $T$ and vehicle capacity $Q$) of POMO on \emph{Distance} PDSTSP42 with a randomly generated instance.}
\label{fig:cosntraint_generalization}
\end{figure}

We also assess POMO's zero-shot ODD robustness to constraint tightness by
varying $T$ and $Q$ to values not have been seen during training. As shown by
Fig.~\ref{fig:cosntraint_generalization}, POMO maintains competitive
performance across a wide range of $T$ and $Q$; since the best-achievable
revenue should be monotone nondecreasing as constraints loosen, POMO's curve
closely tracks this trend with only minor deviations, indicating reliable
adaptation to changing operating conditions without retraining. This robustness
is crucial for real-world applications where constraints may vary frequently.

\subsubsection{Different revenue settings}
\label{appendix:revenue_settings_results} In this section, we evaluate each
method under three revenue settings \emph{Ton-Distance}, \emph{Constant}, and
\emph{Uniform}—to study how payoff structure shapes routing behavior and
relative performance. Across settings, we use the same instance generation,
vehicle capacity $Q$, maximum route length $T$, and (when needed) normalize
revenues to comparable scales. 
From Tab.~\ref{tab:revenue_setting_results}, we observe that the relative
performance of different algorithms remains consistent across revenue settings.
This provides further support for the observations and conclusions drawn in the
main text based on the distance setting. Hybrid methods that combine NCO with
search (e.g., POMO+MSLNS and POMO+BI-LNS) consistently outperform pure NCO (e.g., POMO) and metaheuristics. POMO+MSLNS$_{[10,5]}^{t_{max}}$ achieves
near-optimal performance across all settings, but generally requires more time.

\begin{table*}[t]
\centering
\caption{Performance matrix: PDSTSP42 vs.\ PDSTSP82 across different revenue setting (\emph{Ton-Distance/Constant/Uniform}) with $Q$ and $T$ randomly generated. }
\small
\setlength{\tabcolsep}{3pt}
\resizebox{0.9\textwidth}{!}{%
\begin{tabular*}{\textwidth}{@{\extracolsep{\fill}}
    l      
    l      
    r r r r r   
    r r r r r   
@{}}
\toprule
 & & \multicolumn{5}{c}{\textbf{PDSTSP42}} & \multicolumn{5}{c}{\textbf{PDSTSP82}} \\
\cmidrule(lr){3-7}\cmidrule(lr){8-12}
\multicolumn{1}{c}{\textbf{}} & \multicolumn{1}{c}{\textbf{Algorithm}} &
\multicolumn{1}{c}{Time} & \multicolumn{1}{c}{Revenue} & \multicolumn{1}{c}{Profit} & \multicolumn{1}{c}{Gap} & \multicolumn{1}{c}{WR} &
\multicolumn{1}{c}{Time} & \multicolumn{1}{c}{Revenue} & \multicolumn{1}{c}{Profit} & \multicolumn{1}{c}{Gap} & \multicolumn{1}{c}{WR} \\
\midrule

\multirow{7}{*}{\vlabel{\textbf{Ton-Distance}}}
& GS+BNS$^{t_{max}}$      & 10m & 16.224 & 12.275 & 10.81\% & 16.4\% & 1.2h & 35.459 & 27.933 & 16.77\% & 2.5\% \\
& GS+LNS$^{t_{max}}$     & 17m & 16.085 & 12.134 & 11.58\% & 16.9\% & 1.2h & 35.897 & 28.372 & 15.74\% & 4.6\% \\       
& GS+SA$^{t_{max}}$      & 20m & 16.112 & 12.169 & 11.43\% & 25.7\% & 1.3h & 34.541 & 27.022 & 18.93\% & 4.9\% \\       
& POMO-single            & 37s & 16.060 & 12.207 & 11.72\% & 5.0\% & 4m & 38.806 & 31.434 & 8.91\% & 1.1\% \\
& POMO                   & 6m & 17.136 & 13.253 & 5.80\% & 15.7\% & 1.1h & 40.725 & 33.328 & 4.41\% & 4.5\% \\
& POMO+BNS$^{t_{max}}$    & 15m & 17.844 & 13.912 & 1.91\% & 49.2\% & 2.5h & 41.979 & 34.520 & 1.47\% & \textbf{32.0\%} \\
& POMO+MNS$_{[10,5]}^{t_{max}}$          & 30m & \textbf{17.943} & \textbf{14.029} & \textbf{1.37\%} & \textbf{54.7\%} & 3.3h & \textbf{42.084} & \textbf{34.674} & \textbf{1.22\%} & 30.8\%  \\
& POMO+MNS$_{[10,10]}$   & 1.9h & 18.191 & 14.271 & 0.00\% & 85.1\% & 16.4h & 42.604 & 35.189 & 0.00\% & 81.0\% \\
\midrule
  
\multirow{7}{*}{\vlabel{\textbf{Uniform}}}
& GS+BNS$^{t_{max}}$      & 13m  & 5.434 & 1.443 & 11.75\% & 12.4\% & 1.2h & 11.165 & 3.960 & 16.91\% & 3.1\% \\
& GS+LNS$^{t_{max}}$     & 19m  & 5.342 & 1.346 & 13.24\% & 13.6\% & 1.3h & 11.060 & 3.853 & 17.69\% & 2.5\% \\       
& GS+SA$^{t_{max}}$      & 21m  & 5.395 & 1.407 & 12.38\% & 19.1\% & 1.5h & 10.889 & 3.690 & 18.97\% & 3.9\% \\       
& POMO-single            & 46s  & 5.478 & 1.616 & 11.04\% & 5.6\%  & 4m   & 12.207 & 5.158 & 9.15\%  & 0.8\% \\
& POMO                   & 7m   & 5.840 & 1.934 & 5.16\%  & 16.5\% & 1.3h & 12.869 & 5.790 & 4.24\%  & 4.0\% \\
& POMO+BNS$^{t_{max}}$    & 18m  & 6.054 & 2.078 & 1.68\%  & 53.9\% & 2.8h & 13.259 & 6.092 & 1.33\%  & \textbf{37.1\%} \\
& POMO+MNS$_{[10,5]}^{t_{max}}$ & 34m & \textbf{6.083} & \textbf{2.135} & \textbf{1.20\%} & \textbf{57.1\%} & 3.6h & \textbf{13.303} & \textbf{6.192} & \textbf{1.00\%} & 36.7\% \\
& POMO+MNS$_{[10,10]}$   & 2.2h & 6.157 & 2.198 & 0.00\%  & 86.1\% & 18.1h & 13.438 & 6.323 & 0.00\% & 80.4\% \\
\midrule

\multirow{7}{*}{\vlabel{\textbf{Constant}}}
& GS+BNS$^{t_{max}}$       & 11m  & 8.262 & 4.283 & 16.16\% & 19.9\%  & 1.1h & 16.798 & 9.772 & 22.07\% & 4.2\% \\
& GS+LNS$^{t_{max}}$      & 19m  & 8.168 & 4.192 & 17.12\% & 19.8\%  & 1.4h & 16.627 & 9.598 & 22.87\% & 4.1\% \\       
& GS+SA$^{t_{max}}$       & 21m  & 8.571 & 4.584 & 13.03\% & 32.8\%  & 1.5h & 17.123 & 10.101 & 20.57\% & 7.1\% \\       
& POMO-single             & 47s  & 8.787 & 4.877 & 10.84\% & 26.4\%  & 5m & 19.729 & 12.825 & 8.48\% & 9.6\%\\
& POMO                    & 7m   & 9.418 & 5.482 & 4.43\%  & 56.8\%  & 1.5h & 20.795 & 13.875 & 3.53\% & 33.8\% \\
& POMO+BNS$^{t_{max}}$     & 17m  & 9.643 & 5.730 & 2.15\%  & 76.5\%  & 2.7h & 21.286 & 14.390 & 1.25\% & 72.7\% \\
& POMO+MNS$_{[10,5]}^{t_{max}}$   & 34m  & \textbf{9.786} & \textbf{5.855} & \textbf{0.70\%} & \textbf{90.2\%} & 3.9h & \textbf{21.437} & \textbf{14.544} & \textbf{0.55\%} & \textbf{85.8\%} \\
& POMO+MNS$_{[10,10]}$    & 2.3h & 9.855 & 5.916 & 0.00\%  & 97.0\%  & 19.8h & 21.556 & 14.653 & 0.00\% & 97.2\% \\
\bottomrule
\end{tabular*}}
\label{tab:revenue_setting_results}
\end{table*}

\subsection{POMO vs. Attention Model}

\begin{figure}
\centering
\begin{subfigure}{0.48\textwidth}
    \includegraphics[width=\textwidth]{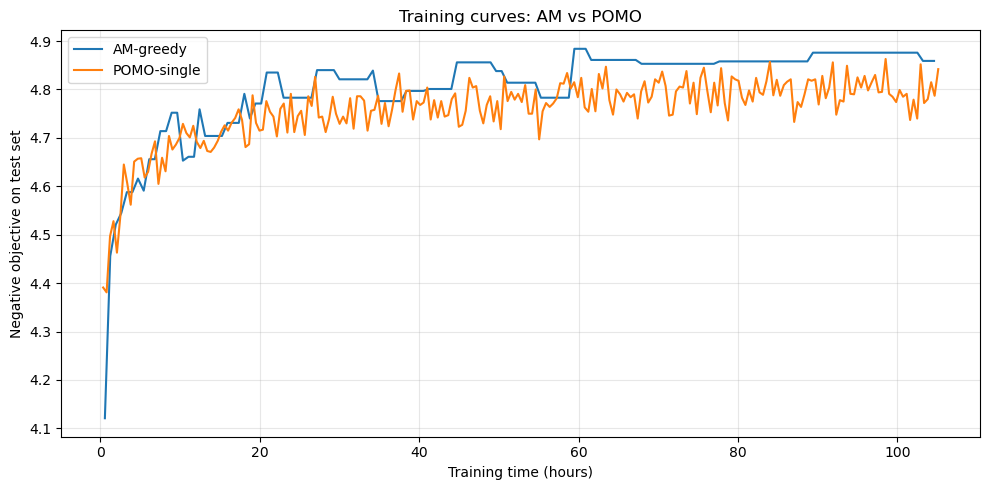}
    \caption{AM v.s. POMO on PDSTSP42}
\end{subfigure}
\begin{subfigure}{0.48\textwidth}
    \includegraphics[width=\textwidth]{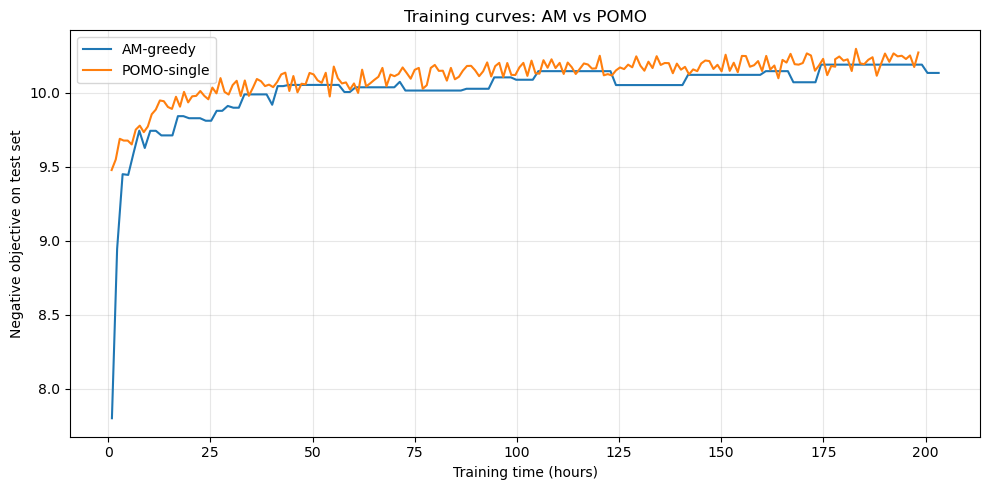}
    \caption{AM v.s. POMO on PDSTSP82}
\end{subfigure}
\caption{Learning curves for \emph{Distance} \textit{m1-PDSTSP} of REINFORCE
with a greedy rollout baseline and POMO. After each training epoch, we generate
100 random instances and employ them as the test set.}
  \label{fig:training_curves}
\end{figure}

From the learning curves in Fig.~\ref{fig:training_curves}, we see the same
overall pattern on both sizes: most of the gain comes early (first ~5-15 hours),
after which improvements are smaller and the curves plateau. For PDSTSP42, the
AM baseline climbs quickly and keeps a small but consistent edge over
POMO-single near the end, whereas for PDSTSP82, the POMO-single continues to
make small gains and slightly surpasses AM. POMO's line is jaggier because we
log every checkpoint; the AM line looks piecewise flat because we evaluate a
fixed baseline that is not updated at every time point. Overall, AM is more
competitive at the smaller scale, while POMO marginally overtakes
as the problem size grows.

\end{document}